\documentclass[journal,transmag]{IEEEtran}
\usepackage{tikz}
\usepackage{makecell}
\usepackage{soul}
\usepackage{cite}
\usepackage{graphicx}
\usepackage{fancyhdr}
\usepackage{xcolor}
\usepackage{lscape}

\usepackage[switch, columnwise]{lineno}
\ifCLASSINFOpdf
\else
\fi

\begin{document}

\graphicspath{{./figs/}}
\DeclareGraphicsExtensions{.pdf,.jpeg,.png, .bmp, .svg}



\title{Applied Exoskeleton Technology: A Comprehensive Review of Physical and Cognitive Human-Robot Interaction}



\author{\IEEEauthorblockN{Farhad Nazari\IEEEauthorrefmark{*},  ~\IEEEmembership{Member,~IEEE},
Navid Mohajer,
Darius Nahavandi, ~\IEEEmembership{Member,~IEEE},\\
Abbas Khosravi, ~\IEEEmembership{Senior,~IEEE}, and
Saeid Nahavandi,~\IEEEmembership{Fellow,~IEEE}}
\IEEEauthorblockA{Institute for Intelligent Systems Research and Innovation (IISRI),
Deakin University, Australia}

\thanks{\IEEEauthorrefmark{*} Corresponding author: F.Nazari (email: f.nazari@deakin.edu.au).}}

\maketitle
\IEEEoverridecommandlockouts
\IEEEpubid{\begin{minipage}{\textwidth}\vspace{50pt}\centering \scriptsize
~\copyright~2023 IEEE. Personal use of this material is permitted. Permission from IEEE must be obtained for all other uses, in any current or future media, including reprinting/republishing this material for advertising or promotional purposes, creating new collective works, for resale or redistribution to servers or lists, or reuse of any copyrighted component of this work in other works. Published in IEEE Transactions on Cognitive and Developmental Systems. DOI: 10.1109/TCDS.2023.3241632
\end{minipage}}







\begin{abstract}
Exoskeletons and orthoses are wearable mobile systems providing mechanical benefits to users. Despite significant improvements in the last decades, the technology is not fully mature to be adopted for strenuous and non-programmed tasks. To accommodate this insufficiency, different aspects of this technology need to be analysed and improved. Numerous studies have tried to address some aspects of exoskeletons, e.g. mechanism design, intent prediction, and control scheme. However, most works have focused on a specific element of design or application without providing a comprehensive review framework. This study aims to analyse and survey the contributing aspects to this technology's improvement and broad adoption. To address this, after introducing assistive devices and exoskeletons, the main design criteria will be investigated from both physical Human-Robot Interaction (HRI) perspectives. In order to establish an intelligent HRI strategy and enable intuitive control for users, cognitive HRI will be investigated after a brief introduction to various approaches to their control strategies. The study will be further developed by outlining several examples of known assistive devices in different categories. And some guidelines for exoskeleton selection and possible mitigation of current limitations will be discussed.
\end{abstract}







\section{Introduction}
\looseness=-1 Assistive technologies can help individuals with a particular task in an environment associated with pain or injury. In the last 30 years, advancements in assistive technology have experienced a significant improvement in design and application within field use. However, most systems still struggle to accommodate slight variations in tasks other than the initially planned application. This can become a critical issue as most strenuous jobs are not singular and often involve different movements.

The concept of exoskeletons goes back to 18\textsuperscript{th} century when Vangestine conceptualised a wearable device for people with disabilities to assist them with walking, jumping and running \cite{Pons2008}. This concept became a reality by the design of the first exoskeleton in 1936 \cite{Bennett1966}. One of the first active devices is the full-body exoskeleton developed by General Electric in 1965 to handle heavy loads \cite{Mosher1967}.

\begin{figure}[t]
	\centering
	\includegraphics[scale=0.5]{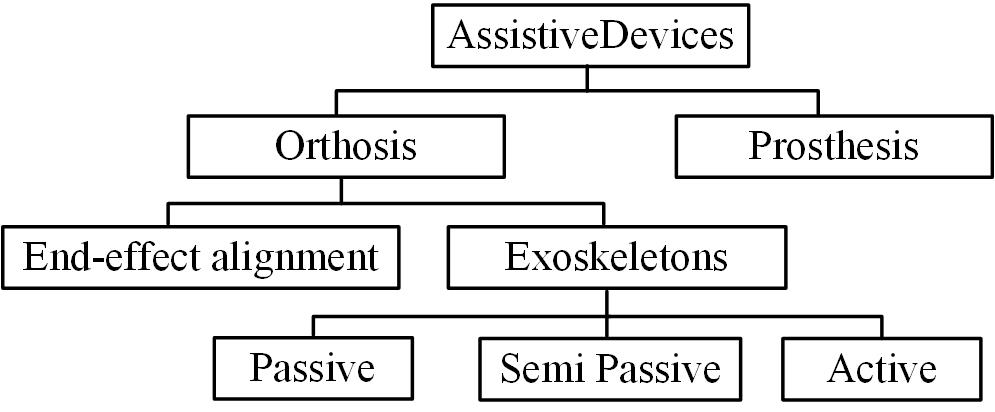}
	\caption{Classification of human assistive devices.}
	\label{fig:cl}
\end{figure}

\looseness=-1 Wearable assistive devices can be generally categorised into prostheses and orthoses. The first one aims to replace a missing body part's functionality, whereas the second one is to support or augment the functionality of the existing part \cite{Gopura2016}. The orthoses can be designed to operate either along the human joints and limbs known as exoskeletons \cite{Sankai2010b, Kazerooni2006} or adjust the user's end-effector \cite{Chen2021a}. A classification of wearable assistive devices has been shown in Fig. \ref{fig:cl}. This survey will be focusing on exoskeletons.

Exoskeletons can also be divided into different groups based on their application, whether they are powered or not \cite{8966474}, the target body part they are assisting with \cite{9340402} as well as their design type, control system \cite{8906035}, user interface \cite{8967017}, and mobility \cite{Sarajchi2021}. These variations may require different assistive strategies. These strategies can affect Human-Robot Interaction (HRI), both at the physical and cognitive levels. For instance, they can be passive \cite{Huysamen2018}, semi-passive \cite{Grazi2020a} and active \cite{Otten2018} in terms of power and control requirements. Passive systems allow the user to carry loads more than the human user's capability without additional support. This may expose users to a risk of injury when the system is not transferring the load to the ground during movement. Additionally, the rigid design of passive systems may cause loss of mobility, transparency, and discomfort. On the other hand, active systems empower the joints with actuators that can provide extra torque. The added torque allows minimal activation of the muscles and thus reduces muscular fatigue, allowing for prolonged tasks. Active systems can be further categorised based on the number of joins/ limbs they are assisting with. In gait training, for instance, Fernandez et al reported that 36\% of exoskeletons use only a single active joint, 52\%  two and 12\% utilise three active joints \cite{Fernandez2021aa}

\looseness=-1 Exoskeletons have shown many advantages over traditional wheeled devices like wheelchairs. They allow the user to perform tasks outside paved surfaces and are usually more transparent and intuitive to control as they follow the user's motion \cite{Pazzaglia2016}. Moreover, they can provide the user with haptic feedback \cite{Pratt2002, Amirpour2022}. They can be designed with different applications in mind, e.g. rehabilitation \cite{Lyu2019}, capability augmentation \cite{Kazerooni2005}, disability support \cite{Diego2021}, helping factory workers \cite{Maurice2019}.

A significant limitation of wearable assistive devices is their energy consumption. Wheeled devices mainly require energy for the initial acceleration and friction compensation. In contrast, exoskeletons and orthoses require energy for constant acceleration and deceleration of the limbs in addition to the energy needed for supporting against gravity \cite{VandenBogert2003}. This is especially important in portable devices and limits their operation range \cite{Young2017}. Various strategies have been investigated to mitigate this issue. These solutions range from reducing the exoskeleton's weight and inertia \cite{Lee2018} to saving the deceleration energy in a spring \cite{Hidayah2021} or an elastic band \cite{Bajaj2021}. A more detailed explanation of the different approaches will be discussed in section II, followed by more examples in section V.


\looseness=-1 The review of published surveys has shown that most of them report either a specific type of exoskeleton, e.g. active lower-limb exoskeletons \cite{Sun2022}, or a particular subsystem of these devices, e.g. control \cite{Kiguchi2007}, pattern recognition \cite{Young2014} classification \cite{Kang2022}. The current survey investigates the design background and development of state-of-the-art exoskeleton technology regardless of their intended use \cite{Toxiri2019, Li2018, Bosch2016, Wehner2013}, utilised technology \cite{Jali2016}, or target body parts \cite{Li2018, Kang2022, Sun2022}. More specifically, assistive technology comprises different levels of physical and cognitive Human-Robot Interaction (HRI). As part of this research, various realisations of HRI with both simple and intelligent implementation will be investigated and several examples will be provided to support the discussion on the design, technical aspects, and limitations mitigation for HRI.

\begin{figure*}[t]
	\centering
 \hspace*{-7mm}  
	\includegraphics[scale=0.83]{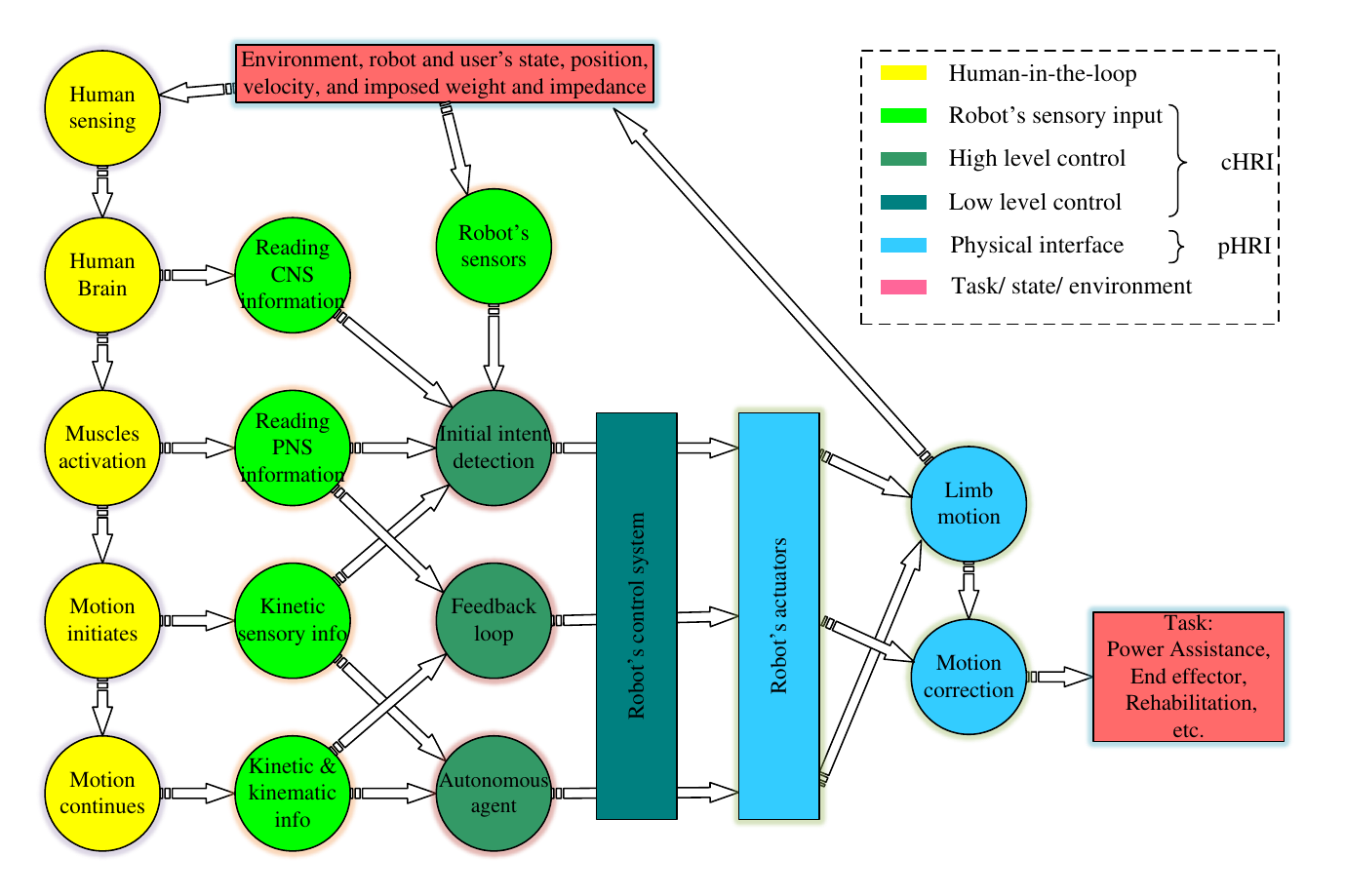}
	\caption{The diagram of the human-robot interaction of a wearable assistive device.}\label{hmi}
\end{figure*}

Human-robot interaction is "a field of study dedicated to understanding, designing, and evaluating robotic systems for use by or with humans" \cite{Goodrich2007}. This concept was a breakthrough in the development of active exosystems \cite{Kazerooni1990}. The interaction between the exoskeleton and the user is either physical or cognitive \cite{Gopura2016}. Physical Human-Robot Interaction (pHRI) determines the suitability of the exosystem, whereas cognitive Human-Robot Interaction (cHRI) consider intelligence for controlling the system \cite{Pons2008}. pHRI, which includes Degree of freedom, actuation, kinematic, transmission, and dexterity, directly affects safety and comfort aspects \cite{Schiele2008}. In contrast, cHRI is crucial for intuitive and real-time control of the system. Fig. \ref{hmi} shows the HRI structure for an active exoskeleton. The bidirectional collaboration between the robot and the human allows for performing shared tasks. The robot receives the human movement intention and, in return, provides them with mechanical help as well as feedback information \cite{Yan2015}. While the robot's output is the outcome of mechanical design orpHRI, the intent interpretation should be implemented at the controller level or cHRI.

The rest of this paper is organised as follows: In Section II, the biomechanics aspects of exoskeletons are explored, and the mechanical solutions, including the mechanism, actuation and energy source are discussed. Section III introduces the control strategies integrating intent detection and interpretation components to provide an intuitive performance for the exoskeleton. In Section IV, the cHRI for wearable assistive devices is investigated. This section discusses intent detection and the relevant subsystems, including sensors, signal processing, data fusion, and different approaches for intent interpretation. In Section V, the most well-known existing systems are classified into passive, semi-passive and active systems, and a detailed discussion of each system is provided. Section VI outlines the technologies covered in previous sections and proposes some guidelines for exoskeleton selection. Finally, Section VII concludes this study.

\section{Physical Human Robot Interaction}~\label{sec:design}
The design and implementation approaches of exoskeleton systems converge despite the difference in their applications \cite{Young2017}. These systems should be aligned with human biomechanics \cite{Pons2008}, wearable and portable \cite{Kim2019a}. Moreover, the power consumption, control subsystem, accuracy and robustness of them play a key role in the design requirement \cite{Desplenter2020}. Overall there are several aspects associated with the design criteria of exoskeleton systems. These include structural and material design, input and output hardware, including sensors and actuators, energy sources, and software or control systems. This section discusses the physical human-robot interaction design criteria of these devices. These include biomechanics, mechanism, actuators and energy sources.

\subsection{Biomechanics}
\begin{figure}[t]
	\centering
	\includegraphics[scale=0.26]{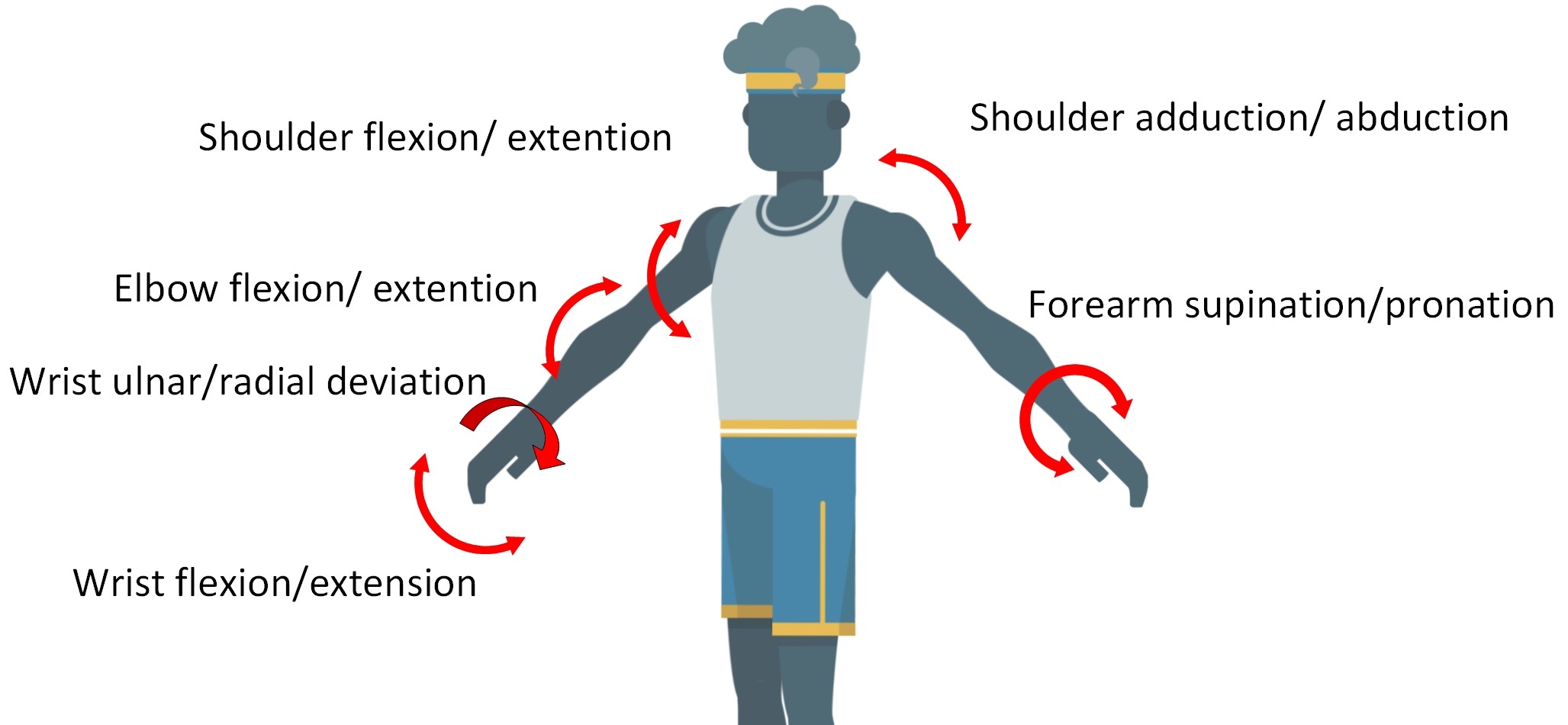}
	\caption{Human upper-limb motions.}
	\label{fig:motion}
\end{figure}

The exoskeleton should be designed around the human body's biomechanics, i.e. anthropomorphism and ergonomics. To make them fully compatible with the human body, the degree(s) of freedom (DoF) \cite{Yang2008} and the placement of actuators \cite{Zoss2006} of the exoskeleton should be aligned with human anatomy. This means that simple yet accurate human kinematic models are essential to developing an efficient control algorithm for these assistive devices. Such a model should involve the DoF of the assisting joints and all possible body parts' movements and their limitations. Fig. \ref{fig:motion} shows the possible motions of human upper limbs. Modelling the accurate functions of human upper limbs can be complicated; therefore, usually, a set of pre-determined movements will be used \cite{Geng2016,Gregory2019,Phinyomark2018}. These models can be simpler for the lower limb as there are 3 DoF in the hip and ankle and 1 DoF in the knees. 

These models shall be aligned with the appropriate biomechanical parameters to accommodate the body parts' kinetics and kinematics. This is also necessary for understanding the power requirement of the active systems. For instance, Dollar and Herr showed that the power at the hip, knee and ankle in level walking is evenly distributed in both the positive and negative sides \cite{Dollar2008}. Another example is weight-climbing, which presents more significant biomechanical challenges as the body's centre of mass (CoM) needs to be raised \cite{Li2018}. Other than that, these parameters help with studying the exoskeleton's impact on the user. There are several ways to measure the effectiveness of an exoskeleton, e.g. average maximum load \cite{Li2018}, joint moment \cite{VanDijk2014}, muscle torque reduction \cite{VandenBogert2003,VanDijk2014}, muscle activity \cite{VanEngelhoven2019,Hull2020,Kobayashi2009} and metabolic cost \cite{Asbeck2013,Collins2015, Grabowski2009,Walsh2007}. Table \ref{tbl:bio} shows some common ways of measuring the biomechanical effect of a wearable assistive device on the user.

\subsection{Mechanism}
The main mechanical criteria in the design of an exoskeleton are: ergonomics and comfort, manoeuvrability, weight, structural strength, adaptability and safety \cite{Onen2014}. The importance of these criteria differs depending on the application and focused limb, affecting the structural design and optimal solution \cite{Gopura2016}. Moreover, an exoskeleton needs to be adaptable to the wearer's characteristics, e.g. body shape and limb size \cite{Hong2013}. Also, it has to deal with the changes in the body motion's characteristics. This is essential to minimise the misalignment between the human and robot's joints and limbs. However, achieving this alignment is challenging as biological joints' centre of rotation are not fixed \cite{Simon1993}. For example, the instant centre of rotation (ICR) at the shoulder and elbow alters with joint motion \cite{Schiele2008}. To solve this problem, MEDARM uses additional joints so it can move the centre of rotation \cite{Ball2007}. Another approach is to consider the shift of the centre of the rotation in the design. An example of this approach is the upper limb exoskeleton by Kiguchi and Fuku \cite{Kiguchi2007}.
\par
\looseness=-1 Misalignment can also occur during motion between the human and robot's joint axis. The resulting high contact pressure and cognitive loads can cause discomfort for the user \cite{Esmaeili2011}. Jarrasse and Morel proposed a hyperstaticity solution to prevent these undesired interaction forces \cite{Jarrasse2011}. Vitiello et al. avoided this misalignment issue in NEUROExos using a compliant actuation system \cite{Vitiello2013}. Furthermore, self-aligning systems like decoupling joint rotation \cite{Stienen2009} and ASSISTON-SE \cite{ErginMehmetAlperandPatoglu2012} are alternative approaches to aligning the robot and user's joints. However, these solutions come with the detriment of added mass and complexity. That is why most systems focus only on one joint, like knee \cite{Pratt2004,Shamaei2014} or ankle \cite{Collins2015}. Table \ref{tbl:mslgn} shows some solution approaches for misalignment in exoskeletons.
\par
Structural integrity or strength is the other aspect of mechanical design. Material property and mechanism design are the key influencers of structural integrity. They must utilise materials with a high strength-to-weight ratio to maximise the system's movability. Light metal alloys like Aluminium have been widely used in many designs like HAL-3 \cite{Young2017}. Titanium has a 67\% better strength-to-weight ratio than Aluminium \cite{Youssef2014}. Fibre-reinforced composites have gotten popular in recent years due to their high strength, lightweight, and ability to shape different designs \cite{Asbeck2013,Kim2019a}. However, they cannot efficiently perform compression and connection strength.


\looseness=-1 The most common exoskeleton designs are rigid robotic mechanisms designed around the human body. They connect to the user's limbs and help them parallel to their muscles and joints. PKAExo \cite{Li2018}, H-PULSE \cite{Grazi2020a}, BLEEX \cite{Zoss2006}, RoboKnee \cite{Pratt2004} and HAL \cite{Sankai2010b} are examples of a rigid design. Their downsides are the added bulk and weight, which may undermine the ease of application and increases energy consumption. For lower limb devices, more actuation energy is required by moving the mass towards the foot and ankle \cite{Browning}. For example, adding 4 kg mass to the centre of mass of a person will increase their metabolic cost by 7.6\%, whereas this number can rise to 34\% if the mass moves towards the feet \cite{Asbeck2013}. One possible approach to prevent this added mass is to use passive systems that store and release kinetic energy in elastic parts \cite{Collins2015,Grabowski2009,Li2018}.

An alternative solution is soft designs inspired by invertebrates \cite{Rus2015}. This design solution can assist the user through flexible structures \cite{Wehner2013} and increases the system's flexibility. The lightweight elastic fabrics used in these systems help reduce inertia and power consumption and increase comfort and transparency. An emerging field in soft systems is soft pneumatic gripper technology. Kai et al. developed a soft hand exoskeleton based on variable stiffness actuators embedded in a glove \cite{Yap2015}. The proposed system is lightweight and easy to use, and the variable stiffness would allow for different finger profiles and movements. Another benefit of these designs is that precise control becomes less of a priority due to the forgiving nature of the system. In cable-driven soft exoskeletons, torque can be provided on multiple joints simultaneously. This reduces the number of actuators and, in turn, weight and efficiency \cite{Kim2019a, Lee2018}. Another example is the soft exosuit developed in WYSS institute that helps with ankle moment \cite{Asbeck2013}. The proposed design does not have rigid parts to help with compression, so human muscle and bone support the load, which may limit the load-carrying capacity.

\begin{table*}[th!]
	\centering
	\footnotesize
	\caption{The biological effect of exoskeleton assist with specific tasks} \label{tbl:bio}
	\begin{tabular}{p{5.7cm} p{1.8cm} p{1.8cm} p{5.7cm}}
		\hline \vspace{1mm}
		
		Ankle exoskeleton \cite{Galle2017}:
		
		- Uses artificial pneumatic muscles to help with ankle push-off.
		
		- Reduce walking metabolic cost by up to 21\%. 
		& \centering
		\raisebox{-\totalheight}{\includegraphics[scale=0.25]{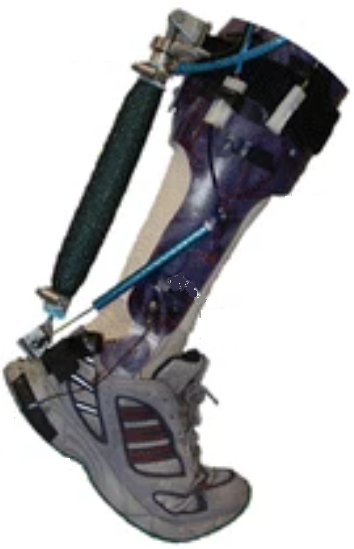}}
		& \centering
		\raisebox{-\totalheight}{\includegraphics[scale=0.1]{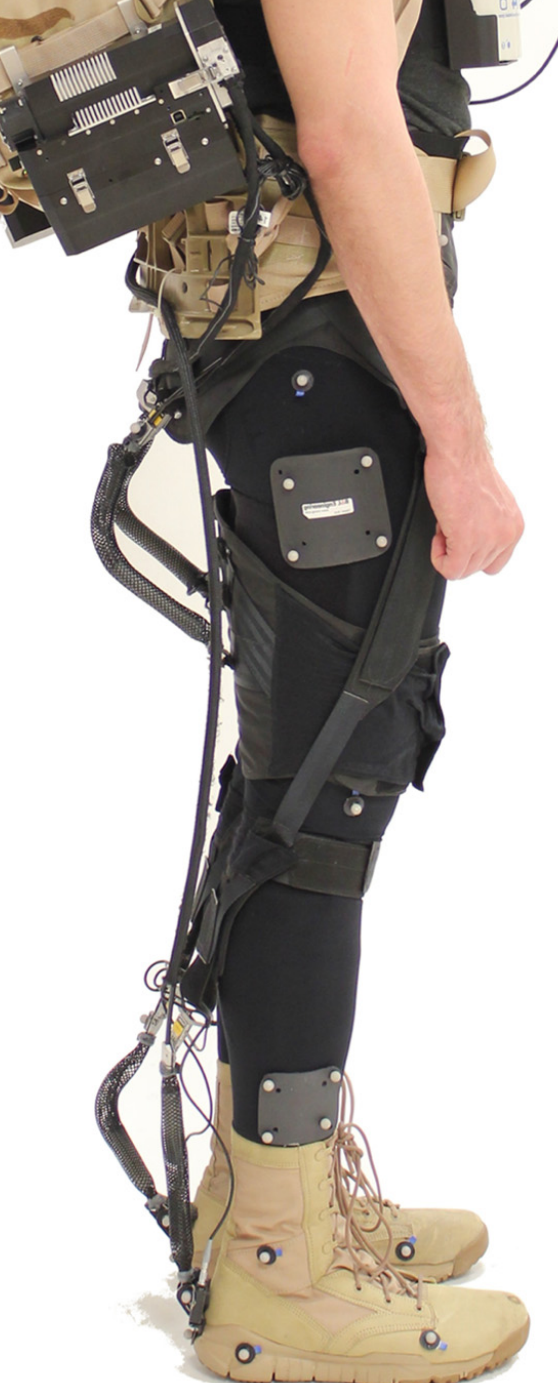}}  
		& \vspace{1mm}
		
		Multi-joint soft exosuit from Wyss Ins. \cite{Panizzolo2016}: 
		
		- Reduces the muscle activity of the vastus lateralis and soleus by 4.7 and 8.4\% during walking
		
		- 14.2\% reduction in metaboloc cost.
		\\ \vspace{1mm}
		
		Passive-elastic knee-ankle exoskeleton \cite{Etenzi2020}:
		
		- Stores knee extension energy in a spring and releases it during ankle push-off. 
		
		- Reduce the metabolic cost by 11\% compared to wearing disengaged exoskeleton.
		& \centering
		\raisebox{-\totalheight}{\includegraphics[scale=0.22]{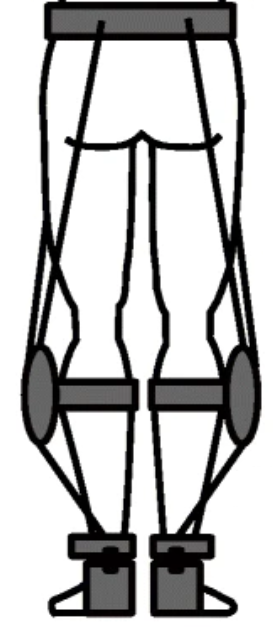}}
		& \centering
		\raisebox{-\totalheight}{\includegraphics[scale=0.6]{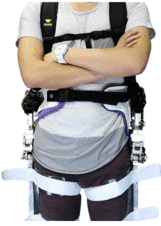}}  
		& \vspace{1mm}
		
		Powered bilateral hip exoskeleton \cite{Kang2019a}:
		
		- Achieved reduction of 6\% in gait energy cost in it's optimum setup. 
		\\ \vspace{1mm}
		
		Autonomous hip exoskeleton \cite{Seo2017}:
		
		- Uses BLDC motor to help hip rotation in uphill walking
		
		- Demonstrated 15.5\% metabolic cost saving at 10 grade slope.
		& \centering
		\raisebox{-\totalheight}{\includegraphics[scale=0.31]{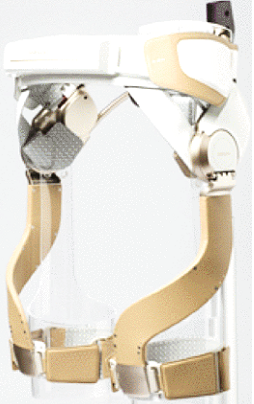}}
		& \centering
		\raisebox{-\totalheight}{\includegraphics[scale=0.12]{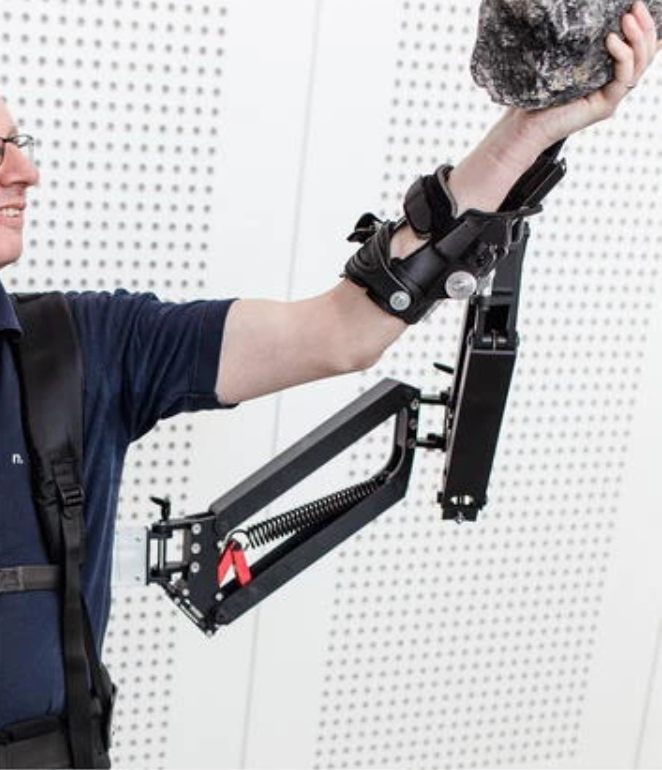}}  
		& \vspace{1mm}
		
		Passive arm exoskeleton \cite{Altenburger2016}:
		
		- Showed 62 and 49\% reduction in Medial Deltoid and Biceps Brachii's muscle activity while holding a 2Kg load \cite{Huysamen2018}.
		\\ \hline
	\end{tabular}
\end{table*}

\subsection{Actuators}
The actuator should provide the required mechanical strength while having minimal weight and transparency. Most of the popular robotic actuators cannot pass the high torque, and high-speed required for exoskeletons \cite{Hong2013}. In general, size, weight, and actuation efficiency can affect the system's range and manoeuvrability.

Among electric actuators, servo motors are best suited for applications that require precise position control. This includes almost all assistive devices aiding disabled or injured body parts, e.g. injured spinal cord \cite{Young2017}. These motors can be set up in direct drive actuator mode, series with reduction gears or cable drive transmissions. While the first approach leads to an expensive, bulky, ungainly set-up, the gear set-up increases friction and reflected inertia. On the other hand, pulley transmission needs a lot of space for large pulley set-ups to get the desired reduction ratio \cite{Pratt2002}.

\looseness=-1 Load cells can be used in closed-loop control systems to avoid friction and inertia problems with geared solutions. The load cell measures the amount of force that the actuator is applying. At the same time, the feedback controller uses the difference between the applied force and the desired one. The downside is that they are susceptible to shocks, e.g. when the foot touches the ground. A solution to this problem is using Series Elastic Actuators (SEA) \cite{Pratt2002}. As shown in Fig. \ref{sea}, SEA works fundamentally like the traditional linear actuators coupled with load cells in a closed-loop system. The difference is that instead of load cells, it uses spring deflection to measure the force. This flexible part can also help with energy-saving, similar to passive devices. High fidelity force control, back-drivability, low impedance and low energy consumption are some advantages.

Brushless dc motors have high torque to weight ratio and can be a perfect candidate for applications that do not require precise position control. 
Ankle exoskeleton by Mooney \cite{Mooney2014} is an example of this type of motor.
In the case of heavy lifting, hydraulic systems could be a suitable replacement for electric and pneumatic actuators \cite{Hunter1991}. Sarcos exoskeleton is an example using hydraulic rotary motors \cite{Hong2013}. However, rotary motors can suffer from friction problems and oil leakage. The linear hydraulic actuators like the ones in the Lower Extremity Exoskeleton (BLEEX) \cite{Zoss2006} can be a proper alternative.

These hydraulic actuators, however, may not be suitable for compact designs \cite{Young2017}. Instead, artificial muscles can be used for the mentioned systems. These artificial muscles can be electroactive, memory alloy or pneumatically controlled. An example of these actuators is flat pneumatic artificial muscles developed by Wyss institute  \cite{Park2014}. Their 2D design and fabric sleeves make them compact and lightweight.

Another novel developed and investigated technology for the actuation of assistive soft robotic suites is Twisted String Actuation (TSA) \cite{Stevens2016}. It works based on two or more strands connected in parallel to an electrical motor, producing linear motion from the rotation of the motor \cite{Palli2013}. These actuators aimed to tackle the weight, size and complexity problems with traditional actuators through their light and compact structure \cite{Popov2013}. Hosseini et al. demonstrated up to 220\% arm muscle activity compensation with an ExoSuit weighting only 1650g \cite{Hosseini2020}.
Table \ref{tbl:actuators} shows the actuator choice in some exoskeletons.

\subsection{Energy source}
\looseness=-1 Exoskeletons require a source of energy that can provide the required range of motion for a specific task duration, yet light enough not to adversely affect the system's mobility. Other than the designs tethered to an external power source \cite{Nassour2021}, the exosystems can use batteries \cite{Arunamithra2022}, small internal combustion engines \cite{Bogue2015}, electrochemical fuel cells \cite{Nasirinezhad2021}, or wireless energy transfer \cite{Drosos2019}. Lithium batteries have relatively high energy-to-weight density, which makes them popular in exoskeletons. With the current technology, most active exoskeletons can function between one to five hours \cite{Young2017}. A solution to this limitation can be regenerative braking, e.g. ankle foot orthosis by Oymagil et al. \cite{MehmetOymagil2007}. Overall, this aspect may impose limitations on exoskeletons and their application. To overcome this, more efficient and advanced energy transfer and storage as well as processing and actuation techniques are required.

\begin{table*}[th!]
	\centering
	\footnotesize
	\caption{Reported solutions to misalignment in exoskeletons} \label{tbl:mslgn}
	\begin{tabular}{p{6.5cm} p{1.8cm} p{2.cm} p{6.1cm}}
		\hline \vspace{1mm}
		
		MEDARM for rehabilitation \cite{Ball2007}:
		
		- Uses additional joints to move the centre of rotation to solve the misalignment issue.
		
		- Improves the replication of the motion for rehabilitation by enabling any level of gravity compensation or movement assistance, preventing compensatory movements, and encouraging stroke patients with any level of motor disability to move more naturally.
		&
		\raisebox{-\totalheight}{\includegraphics[scale=0.19]{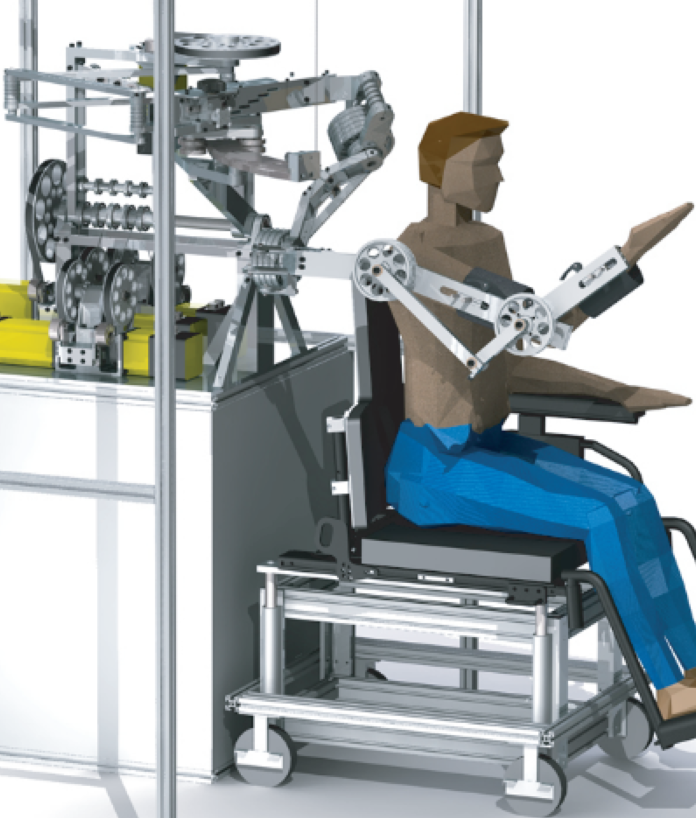}}
		&
		\raisebox{-\totalheight}{\includegraphics[scale=0.32]{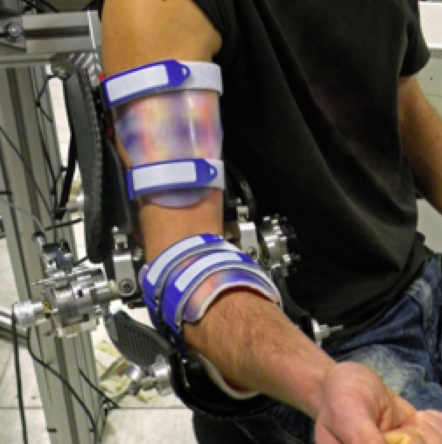}}  
		& \vspace{1mm}
		
		NEUROExos \cite{Vitiello2013}:
		
		- Poststroke elbow rehabilitation exoskeleton.
		
		- Passive mechanism allows the elbow to align with robot's axes during motion.

            - The joint stiffness can be adjusted between zero to 60 N.m/rad, and can be configured in both robot-in-charge and patient-in-charge for rehabilitation.
		\\ \vspace{0.5mm}
		
		SPEXOR \cite{Koopman2020}:
		
		- Revolute-Revolute-Revolute (RRR) \cite{10.3389/frobt.2018.00072} misalignment compensation method for the hip.
		
		- It stores energy during bending and releases it for back extension, and depending on the bending angle, it lowers the compression force by 13-21\%.
		&
		\raisebox{-\totalheight}{\includegraphics[scale=0.32]{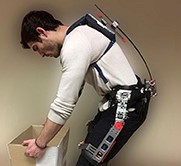}}  
		&
		\raisebox{-\totalheight}{\includegraphics[scale=0.26]{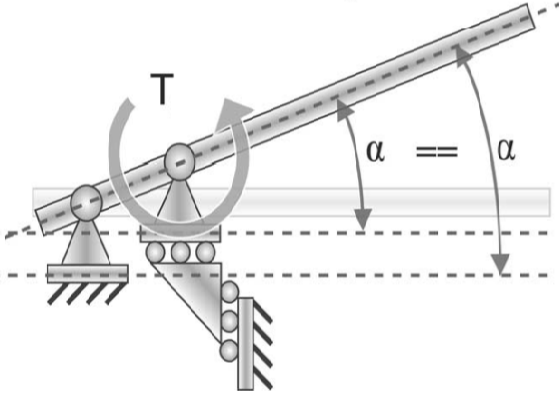}}
		& \vspace{1mm}
		
		- Fixing the misalignment and eliminating contact pressure in exoskeletons by translating axis and preventing the interaction from becoming painful \cite{Stienen2009}.
		
		- The implementation is by using two extra links and the elbow joint.
		\\ \vspace{1mm}
		
		Double-Layered Elbow Exoskeleton \cite{Awad2020}:
		
		- Solves axis misalignment using 3-PRR planar parallel mechanism.
  
		- In their qualitative tests, users ranked usefulness, ease of use, and ease of learning 5, 6.5 and 6.3 out of 7.
		&
		\raisebox{-\totalheight}{\includegraphics[scale=0.16]{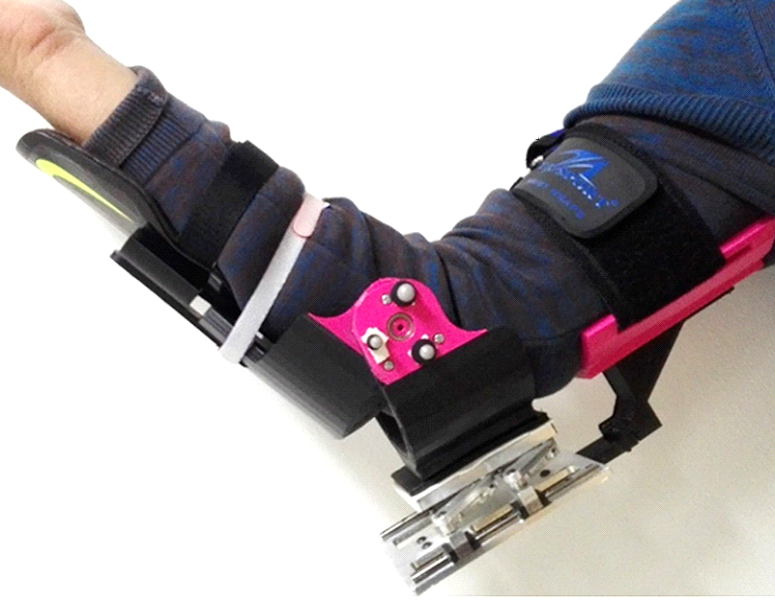}} 
		& \centering \raisebox{-\totalheight}{\includegraphics[scale=0.23]{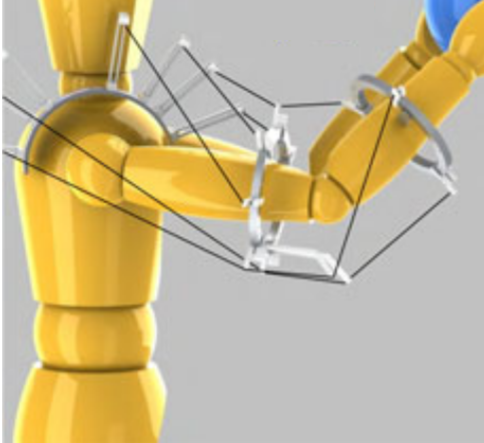}} 
		& 
		
		Cable-Driven Arm Exo (CAREX) \cite{Mao2012}:
		
		- Uses  cable lengths and joint angles to calculate the glenohumeral joint rotation centre to improve achieving desired forces on the hand.
  
		- Subject tests show between 47 to 76.9 $cm^2$ average deviation for a 26 cm task in different sessions.
		
		\\ \vspace{1mm}
		
		Quasi-passive leg exoskeleton \cite{Walsh2007}:
		
		
		 - Hip misalignment compensation with a cam mechanism
  
		- It could transfer 80\% load of a 36 kg payload directly to the ground, as well as a 23\% decrease in metabolic cost. 
  
		& \centering
		\raisebox{-\totalheight}{\includegraphics[scale=0.18]{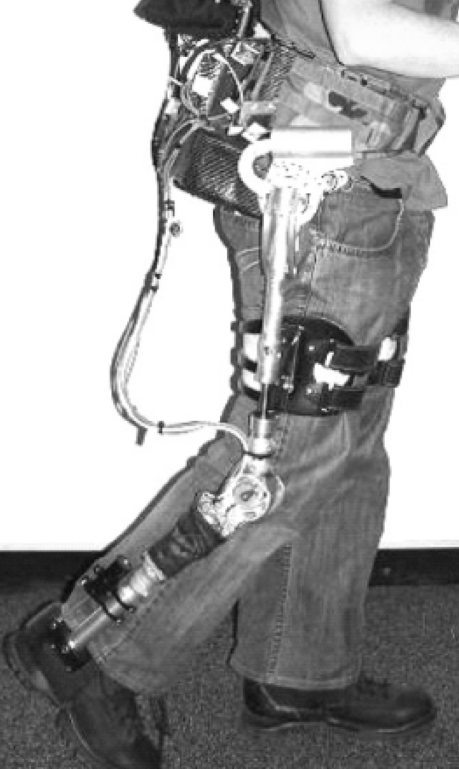}}  
		&  
		\raisebox{-\totalheight}{\includegraphics[scale=0.3]{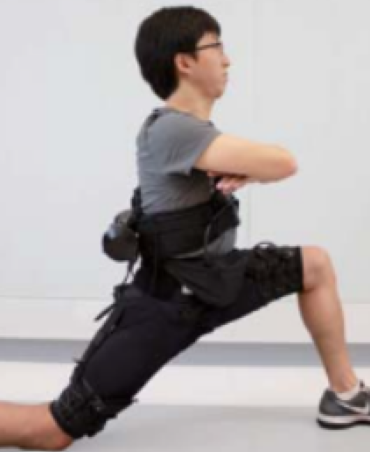}}
		& \vspace{1mm}
		
		Soft exosuit for hip assistance \cite{Kim2019a}:
		
		- Soft material used in the design imposes no restriction on hip movement, solving the problem of misalignment altogether.
  
		- The suit achieved 9.3 and 4.0\% reduction metabolic cost in walking and running.
  
		\\ \hline
	\end{tabular}
\end{table*}

\begin{table*}[tbh!]
	\centering
	\footnotesize
	\caption{Actuators for active exoskeletons} \label{tbl:actuators}
	\begin{tabular}{p{6.3cm} p{1.8cm} p{1.8cm} p{6.3cm}}
		\hline
		\vspace{1mm}
		
		Sarcos Exoskeleton \cite{Hong2013}:
		
		- Uses rotary hydraulic motors to actuate the joints. Figure from \cite{Bogue2009}.
		
		- Sadly, the details of the design and effectiveness of the Sarcos exoskeleton are publicly available.
		&
		\centering
		\raisebox{-\totalheight}{\includegraphics[scale=0.175]{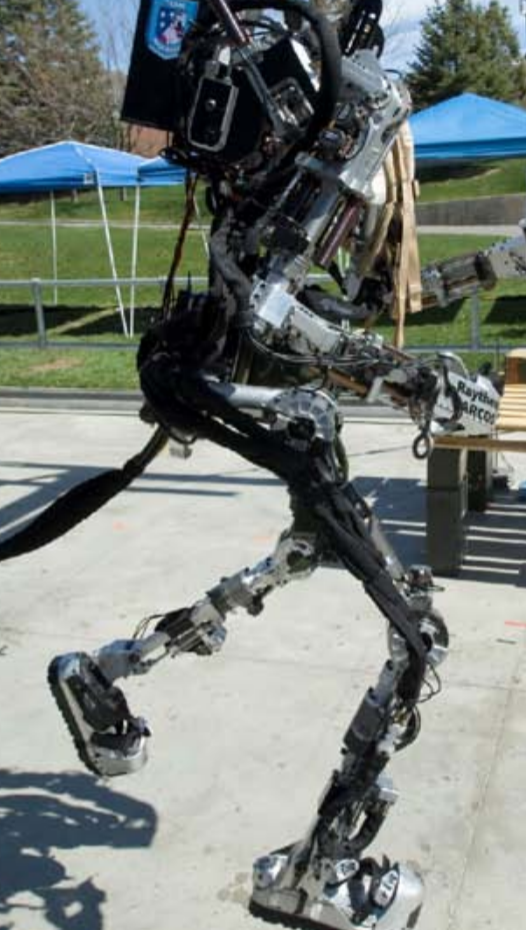}}
		&
		\centering
		\raisebox{-\totalheight}{\includegraphics[scale=0.11]{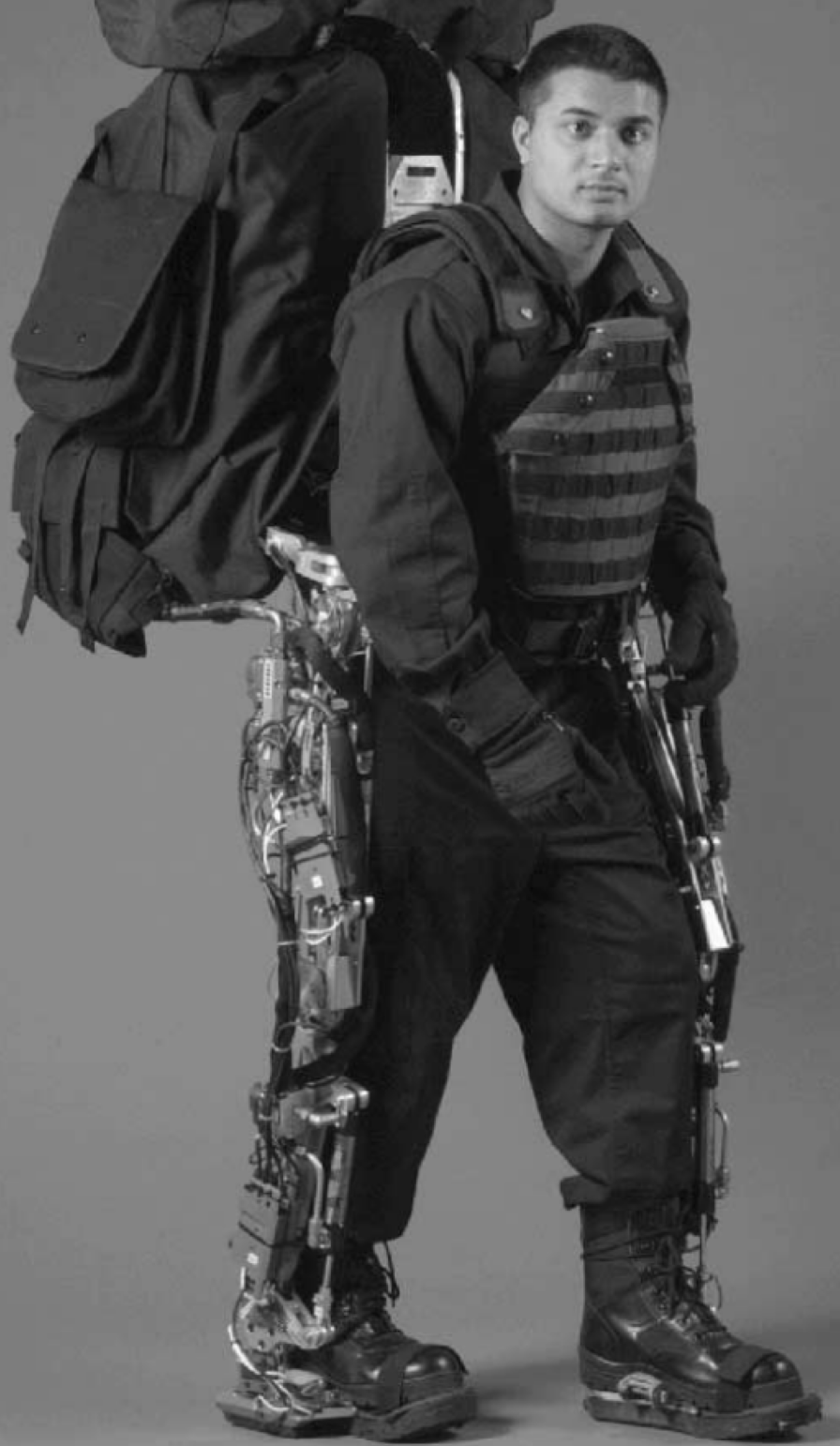}}
		& \vspace{1mm}
		
		Berkeley Lower Extremity Exoskeleton (BLEEX) \cite{Zoss2006}:
		
		- Uses linear hydraulic cylinders instead of rotary due to their leakage.
		
		- It allows the user to carry additional 34 kg of load.
		\\ \vspace{1mm}
		
		Elbow Assistive Soft ExoSuit by Hosseini et al. \cite{Hosseini2017}:
		
		- Integrates a Twisted String Actuation (TSA) module with a force sensor to the forearm of the user and limits muscle activation.
		
		- After 0.5 s, it can reduce the muscle activation from $18.6 \mu v$ higher threshold to $14.4 \mu v$ lower threshold.
		\color{black}
 
		& \centering
        \raisebox{-\totalheight}{\includegraphics[scale=0.2]{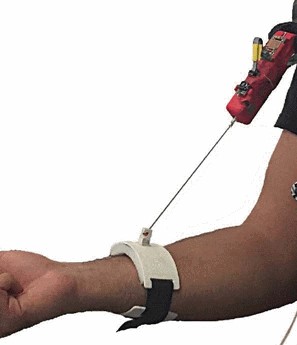}}
		& \centering
		\raisebox{-\totalheight}{\includegraphics[scale=0.53]{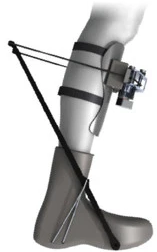}}
		& \vspace{1mm}
		
		Ankle exoskeleton \cite{Mooney2014}:
		
		- Uses brushless dc motor and belt transmission.
		
		- Creates tenstion in the cord connecting to the strut, helping with ankle torque.
		
		- Reduce the walking metabolic cost by 36 W or 8\%.

		\\  \vspace{1mm}
		
		Biologically-inspired soft exosuit \cite{Asbeck2013}:
		
		- Uses Bowden cables to transfer the work from actuation units located on a rucksack on the user's waist to the ankle area.
		
		
		- The suit reduced the peak ankle and hip moments by up to 18 and 30\%.
  
		& \centering
		\raisebox{-\totalheight}{\includegraphics[scale=0.15]{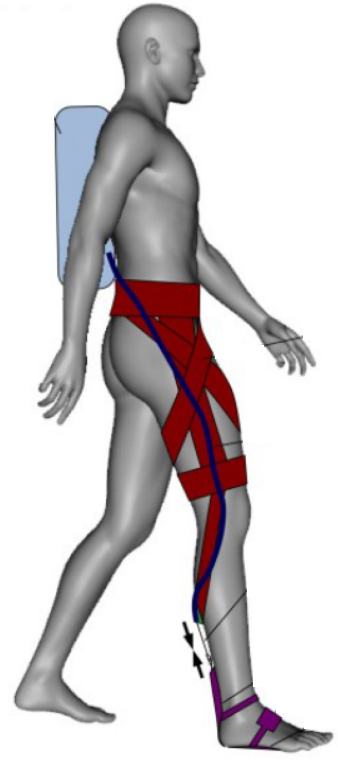}}    
		& \centering
		\raisebox{-\totalheight}{\includegraphics[scale=0.22]{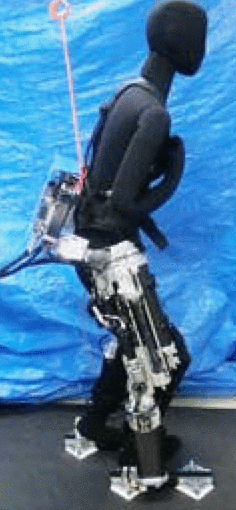}}
		& \vspace{1mm}
		
		ATR hybrid drive eXoskeleton Robot (XoR) \cite{hyon2011}
		
		- Combines a small servo motor with pneumatic artificial muscles (PAM) 
		
		- Hybrid drive takes advantage of the high force and low weight of the PAM and quick response of the servo motor. 
		
		- The robot have 120 deg range of motion and 150 N.m load capacity.
  
		\\  \vspace{1mm}
		
		Muscle suit developed \cite{Kobayashi2009}:
		
		- Utilises McKibben Artificial Muscle. 
		
		- McKibben consists of a braided mesh and an internal bladder which contracts by pressuring the internal bladder. 
		
		- 40\% muscle power reduction achieved in lifting involve sacroiliac.
		& \centering
		\raisebox{-\totalheight}{\includegraphics[scale=0.11]{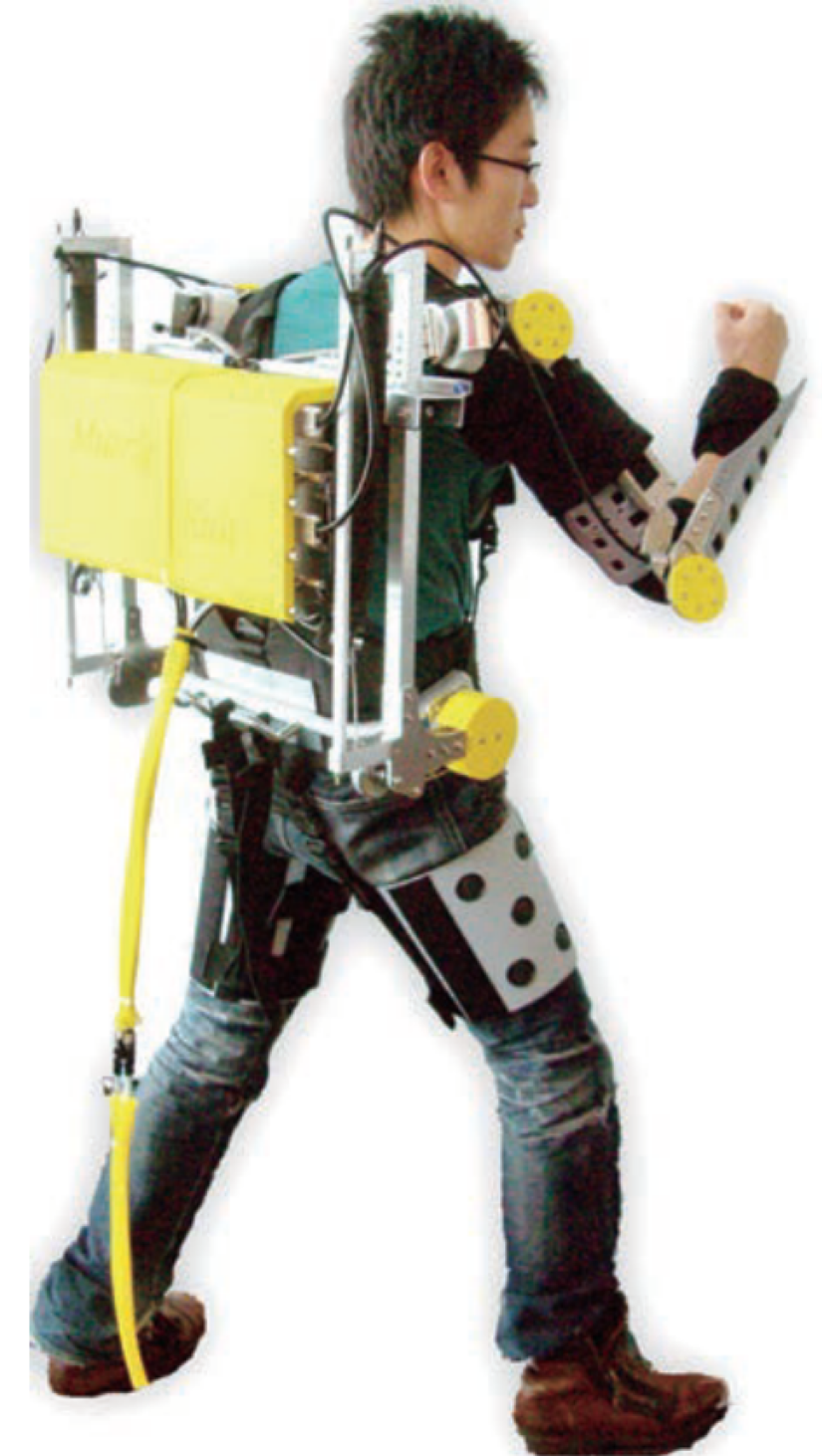}}    
		& \centering
		\raisebox{-\totalheight}{\includegraphics[scale=0.13]{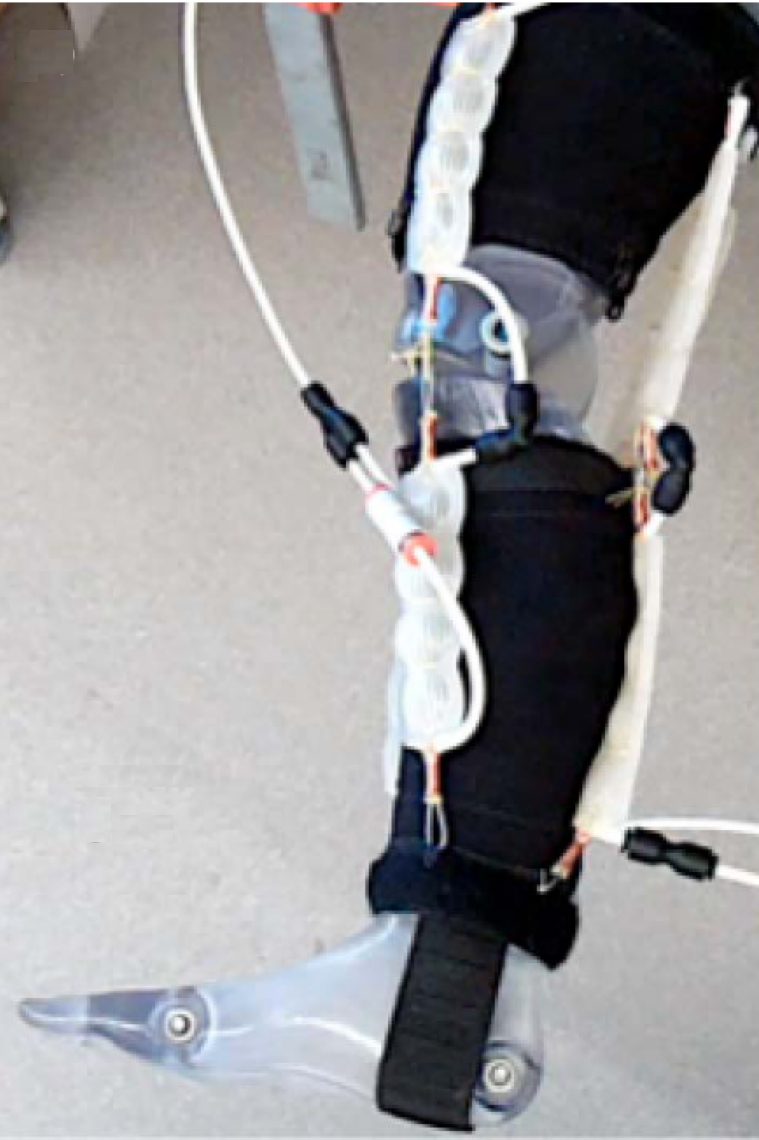}}
		& \vspace{1mm}
		
		Soft knee exoskeleton by Wyss Ins. \cite{Park2014}:
		
		- Uses flat pneumatic artificial muscles. 
		
		- The muscles have 2D elastomeric design that contract with air pressure. 
		
		- It demonstrated 37 and 95$^{\circ}$ knee flexion and extension angle, and 7 and 2.5N force, respectively.
		\\ \vspace{1mm}
		
		RoboKnee \cite{Pratt2004}: 
		
		- 1 DoF knee exo with high transparency. 
		
		- Uses Series Elastic Actuator to achieve low impedance. 
		
		- The user could perform one-legged deep knee bends with a 60 kg backpack load without getting tired.

		&  \centering 
		\raisebox{-\totalheight}{\includegraphics[scale=0.17]{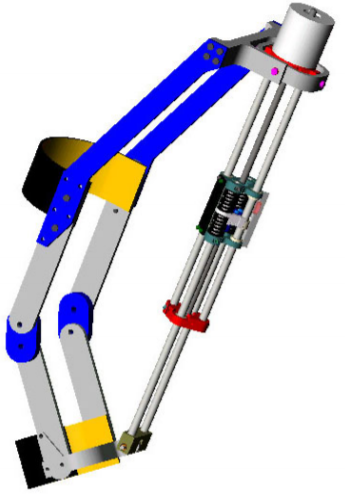}}    
		&  \centering \raisebox{-\totalheight}{\includegraphics[scale=0.16]{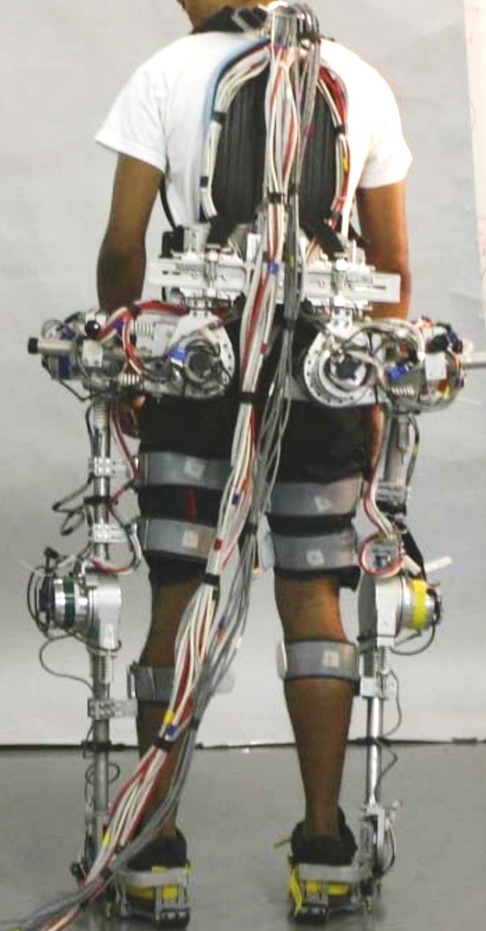}}
		& \vspace{1mm}
		
		IHMC \cite{Kwa2009}:
		
		- 10DoF mobility augmentation exoskeleton. 
		
		- Uses Rotary Series Elastic Actuator (RSEA) to control the hip roll and pitch and knee flexion/extension. 
		
		- It can enable a person who cannot walk without assistance to walk for 15 ft.
		
		\\ \hline
	\end{tabular}
\end{table*}

\section{Control Strategies}
Effective and efficient control systems are necessary to maximise the exoskeletons' performance and achieve stability and safety. The most popular control schemes are kinematic-based and position-based controls \cite{Young2017}. The latter is more beneficial for people with disabilities, i.e. when the user cannot interact with the robot appropriately \cite{Kawamoto2003} by the patient in a set of pre-determined movement trajectories. However, it will decrease the user's manoeuvrability. 

Another categorisation of control systems is whether they are open- or closed-loop. The open-loop (pre-defined/playback) controller applies a pre-determined force/torque, depending on the calculated position. The advantage of this method is that it is easy to embed other variables like step length or walking speed in the gait cycle. However, it is not adaptable to different conditions, e.g. walking on uneven terrain, climbing stairs, or jogging. On the other hand, the closed-loop or feedback system applies force/ torque by processing the input from sensors. Unlike the open-loop ones, these controllers are adaptable to different situations. However, they are more complicated when it comes to implementation \cite{Young2017}. 

There are several approaches or strategies to control assistive devices:

{\it Model-based control:} It is one of the most popular assistive strategies in which the desired joint position or torque will be calculated directly. However, it requires an accurate model of the coupled human-robot dynamic system, which depends on a series of sensors to validate the kinematics and dynamics variables. An example of this control strategy is Hybrid Assistive Limb (HAL) \cite{Sankai2010b}. It uses a model-based strategy to control the knee joint in which electromyography (EMG) signals detect the user’s intention to apply assistive torque, damping or gravity compensation based on the human-exoskeleton model \cite{Kawamoto2010}. Teramae et al. developed a model predictive control (MPC) controller to enhance the rehabilitation of motor muscle functions in patients with torque deficiency in their joints \cite{Teramae2018}.

\looseness=-1 {\it Sensitivity amplification control:} This control technique is based on the inverse dynamic model and enhances the performance of the exoskeleton robotic system and consequently contributes to the load-carrying capability of the human-robot system. The user’s exerted force will be set on a positive feedback loop and can be scaled down, depending on the situation \cite{Kazerooni2005}. Naval Aeronautical Engineering Institute Exoskeleton Suit (NAEIES) \cite{Yang2009} and BLEEX \cite{Zoss2006} are examples of using this strategy.

{\it Predefined gait trajectory control:} \looseness=-1 In this control technique, the joint trajectory is obtained from either gait data analysis or will be recorded from healthy participants. This system aims to help patients with normal voluntary movements. ATLAS \cite{Sanz-Merodio2012}, HAL \cite{Lee2002}, and ReWalk \cite{Esquenazi2012} are examples of this implementation. This assistive strategy seems more straightforward to implement; however, the patient has to walk in an unnatural reference gait.

{\it Adaptive oscillators-based control:} This controller predicts the limb’s future posture based on the periodicity of the gait pattern \cite{Righetti2006}. It captures the phase, frequency, amplitude and offset of the periodic locomotion. This strategy is mainly used in rehabilitative walking and cyclic exercises \cite{Ronsse2011a}.

{\it Fuzzy models:} The adaptive controllers based on fuzzy-logic layers are practical when an accurate dynamic model is hard to construct. The downside is that many parameters shall be tuned according to each motion and person \cite{Ross2010}. To do so, the fuzzy controller needs three main blocks: fuzzification to interpret the inputs; fuzzy-rules block to hold the control knowledge of the system; and defuzzification block to convert the results to the desired output signals. EMG-based neuro-fuzzy controller by Kiguchi is an example of this method \cite{Kiguchi2007}. In another effort, a neuro-fuzzy matrix modifier was used to make an upper-limb exoskeleton adaptable to different users \cite{Kiguchi2012}.

{\it Predefined action-based control:} This technique utilises the recurrent gait phase transition to regulate the actions in flexible interactions \cite{Asbeck2013}. In other words, it works by activating physical impedance and compliance, e.g. spring or pneumatic cylinder and acts in sync with the expected gait task \cite{Walsh2006}.

{\it Hybrid assistive strategy:} This strategy utilises multiple assistive strategies to deal with different situations with increased adaptability. It splits the controller into sub-control states that can kick in depending on the gait cycle phase. For instance, BLEEX's controller is divided into positive feedback sensitivity amplification force and a position controller for swing and stance phase \cite{Kazerooni2006}. Moreover, HAL uses two complementary control algorithms to control the robot based on human intent command and physical body situation \cite{Sankai2010b}. Another example is the hybrid control strategy proposed by Rodriguez et al. for force and end effector position control of a twisted string actuator via two nested loops that monitor the motor angle and the clamping axial force \cite{Rodriguez2021}.

\section{Cognitive Human-Robot Interaction}
 Cognitive HRI plays a key role in the wearer's comfort and safety. This subsystem can vary significantly from one design to another. For instance, Sarcos exoskeleton utilises direct input from the wearer through force sensors interfacing the human body to the robot \cite{Hong2013}. BLEEX, on the other hand, does not attach any sensor to the user; Instead, it relies on accelerometers, encoders and load distribution sensors to calculate the user and robot's state and predict the intention and control the system \cite{Dollar2008}. HAL utilises surface EMG for its Cybernic Voluntary Control along with position-based and force-based sensors for its complementary Autonomous Control \cite{Sankai2010b}. The advantage of EMG is that even patients with weak muscles, as long as they produce neuro-muscular signals, can utilise it as the input to the robot \cite{Fleischer2005}.
 

Human body parts get the intent orders from the central nervous system and provide them with feedback information. The same concept applies to wearable exoskeletons or prostheses. This feedback should include sensory details such as interaction forces and the action and future intention \cite{Losey2018}. The most common modalities of sensory feedback are haptic, aural and visual. The implementation of motion-force coupling in haptic is similar to how the human body utilises the peripheral nervous system to monitor the task and environment and adjust the action force of the muscles \cite{Losey2018}. Haptic feedback can be divided into kinesthetic and tactile feedback. The first one applies force to the muscles and joints, whereas the latter stimulates mechanoreceptors in the skin.

The control structure of active exoskeletons should be designed with human-in-the-loop consideration. Fig. \ref{hmi} shows the structure of human-robot interaction for a wearable assistive device: In the first layer, both the human and machine sense the information from each other and the environment. Next, the information from the previous layer will be processed in the human brain. Then, the user initiates the intended movement with respect to the robot's impedance, which the machine will detect. This intent detection can be extracted from different levels of human intelligence, with the electroencephalogram (EEG) \cite{8239668}, EMG \cite{Foroutannia2022}, and force myography (FMG)  \cite{Cho2016} being examples of intent extraction from different levels.

After detecting the initial intent, the robot activates the actuators to contribute. The actuators' feedback will be sensed by both the human and robot in the first layer to correct the decision and helping ratio in a closed-loop system. This feedback loop correction continues during the motion until task completion. Also, the motion data could be used for autonomous control of the exoskeleton \cite{Kazerooni2006, Kawamoto2010}. In other words, the robot will act based on either the user or the system's command generator, then senses the environment's response and provides the user with feedback to intuitively execute the task.

In the following sub-sections, after reviewing the control division between the human user and autonomous agent, user intent detection systems with signal processing and data fusion methods will be discussed, and artificial intelligence methods for interpreting the input signals will be covered.

\subsection{Arbitration}
\looseness=-1 Arbitration is an integral part of the human-robot interaction that determines how control is divided between the robot and the user \cite{9186613}. There are different types of arbitration. One common approach is to put the human in charge of position control of the end effector, while the robot assists with the orientation \cite{Losey2018}. Alternatively, both human and robot could control the position and orientation with different influence levels. Arbitration can improve the exoskeleton's performance and mitigate the inaccurate intent detection and prioritise autonomous agent when control signals are spars or imperfect \cite{Gopinath2022}. The human and robot roles in an exoskeleton can be co-activity \cite{Varol2010}, master-slave/ primary-replica \cite{Mao2012}, teacher-student \cite{Powell2012}, or collaboration \cite{Li2015}. These roles influence the design of the cHRI depending on the application. For instance, Topini et al. utilised primary-replica and teacher-student strategies to control a hand exoskeleton for rehabilitation purposes \cite{Topini2021}. Li et al. implemented collaborative arbitration in their assimilation control in a dual-arm exoskeleton to better predict the user's near future behaviour and improve the task interaction trajectory \cite{Li2022}.

\begin{figure}[t]
	\centering
	\includegraphics[scale=0.66]{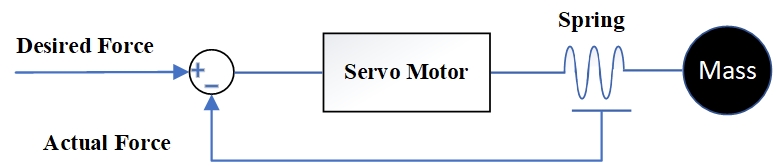}
	\caption{Schematic diagram of Series Elastic Actuator.} \label{sea}
\end{figure}

\subsection{Intent detection}
\looseness=-1 Intent can be defined in many ways depending on the scenario. In some cases, like whether the human wants to control the robot or not, the intention can be defined in binary terms. In controlling a lower limb exoskeleton, they can be defined as a set of possible movements. In continuous systems, intent can be defined as velocity, position trajectory \cite{Pehlivan2016}, force \cite{Pehlivan2016}, or torque \cite{Lenzi2012}. However, it can be more complicated in the upper limb systems, which include more poses and functions. Well defining the intent can help accurately measure the engagement of the patients undergoing neurorehabilitation \cite{Sarac2013}.

\begin{figure*}[t]
	\centering
	\includegraphics[scale=0.55]{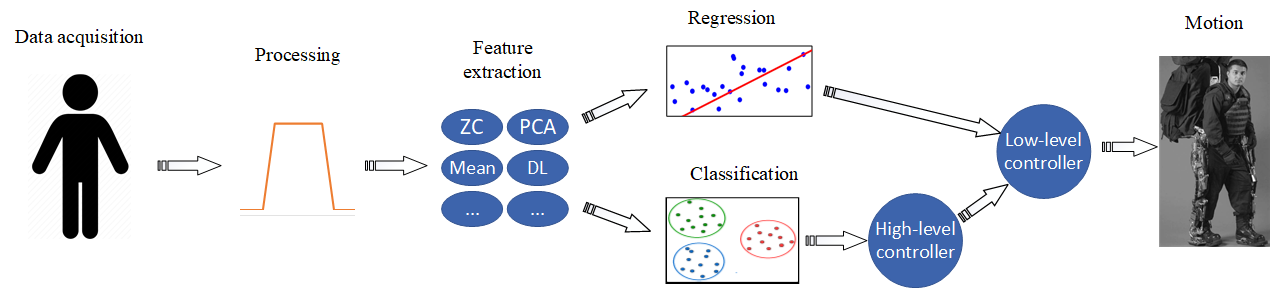}
	\caption{Overall structure of the robot's decision-making process.}\label{dm}
\end{figure*}

Measuring the pre-defined intent can be done in a variety of neural ways like EMG and EEG as well as kinetic- and kinematic-based methods such as FMG, sonomyography, load cells and position encoders to determine the gait phase and assist accordingly \cite{Losey2018}. 
Joint Muscular Torque (JMT) is a bio-mechanical measure of human effort and can be used in the implementation of active power-assist functions. Li et al. developed a novel torque solution to calculate the lower limb's JMT using the inverse dynamics of the body paired with a sensing system \cite{li2018t}.
Another approach to intent detection in mechanical systems is to measure the robot’s position and estimate the indented force or vice versa. Ge et al. developed a method estimating the human intention in the pre-defined limb trajectory by measuring the position, velocity and interaction force of the contact point using neural networks \cite{Ge2011}. They predicted the desired trajectory with $\sim 10e-8$ mean squared error (MSE). However, the phase difference in impedance function causes a time delay between the robot arm and the desired trajectory. In another study, the interaction force in the handles of the intelligent walker is used to forward the path trajectory \cite{Huang2005}. BLEEX minimises the force interaction between the human and robot by shadowing the user’s movement \cite{Kazerooni2005}.

\subsubsection{Sensors}
Sensors play essential roles in assistive devices. Their primary use is to gather data on the operator or patient's kinematics during a task \cite{Baker1976}. They help the exoskeleton to interact with the user and the environment and regulate the output. An example is using position encoders to modulate the torque output of a joint by providing the phase variable \cite{Young2017}. They can also help with safety regulation and calculating the system cycle stride's joint angle trajectory. Another widely used sensors are load cells. They can be highly accurate \cite{Wang2009} and are commonly used in lower limb devices \cite{Schmiedeler2018}. A typical lower limb application measures the contact force between the user and the ground. In some cases, this measurement can be replaced with a binary on/off switch \cite{Young2017}. Series elastic actuators (SEA) can be an alternative to load cells. They measure the deflection of the spring that is installed parallel to the linear actuator and calculate the actuator's force \cite{Pratt2004}. Fig. \ref{sea} shows the schematic of SEA.

In recent years, neural command sensing technologies like EMG \cite{Foroutannia2022} and EEG sensors \cite{Gancet2012} became popular in exoskeletons. These sensors read the user's intention from the peripheral or central nervous system \cite{Nicolelis2001,Velliste2008}. EMG signal is a bio-electric signal caused by the neuro-muscular activity collected directly from the target muscle. The time delay between the neuro-muscular activity and actual movement makes it perfect for real-time applications. For these scenarios, the maximum perceivable delay should be no more than 300 ms \cite{Englehart2003}, which includes this delay, the signal observe period or window size, and the computation time.

Surface EMG sensors collect neuro-muscular electrical signals from the skin surface. As the muscle signal is not collected directly from the targeted muscles, the signal strength will decrease, resulting in a reduced signal-to-noise ratio (SNR). Other than that, the other muscles' activity can affect the quality of the recorded signal, known as cross-talk. If the targeted muscle is hidden behind other muscles, recording a clear muscle activity signal will become harder.

\looseness=-1 In surface EEG, electrodes collect the local field potential (LFP) signals from the scalp non-invasively. This can help to predict the user's intention 500ms before their motion \cite{Lew2012}. In another more invasive approach named electrocorticography or intracranial EEG, electrodes will be surgically placed on the brain's surface to collect the signals with higher precision and less noise \cite{McMullen2014}.

The muscle activity can be alternatively measured without attaching anything directly to the user's body would be FMG. It works by topographically mapping the muscle pressure force \cite{Losey2018}. The muscle force will be calculated based on changes in the muscle volume surface tactile sensors \cite{Cho2016}. The downside is the lack of time-delay benefit of getting the command directly from the central nervous system (CNS). Also, sonomyography can be used to measure the activity of the deep muscles based on ultrasonic technology. Akhlaghi et al. reported a similar accuracy to sEMG in creating forearm muscles' cross-section images using a standard clinical ultrasound system \cite{Akhlaghi2016}. However, this prediction happens $279\pm5$ after motion initiation.

The type of intent measurement directly affects the SNR and determines the system’s robustness. This is especially important in non-invasive bio-signals measurement, where external noises can be far greater than the actual signal amplitude. To avoid system instability, one can use kinetic and kinematic measurements for intent prediction \cite{Losey2018}. However, the time delay between the nervous system and muscle movement may be crucial in some applications. Furthermore, a combination of biosignal and mechanical sensors can be used to enhance the intent prediction \cite{Young2014}. Krausz et al. paired EMG sensors with a gaze recognition system to better coordinate an arm prosthesis \cite{Krausz2020}. This approach has introduced performance benefits by reducing the target estimation root-mean-square error (RMSE) from 9.28 to 6.94 cm while doing that in just 121 ms from the end of gaze fixation.


\begin{table*}
	\footnotesize
	\centering
	\caption{Some approaches for intent prediction} \label{tbl:hmi}
	\begin{tabular}[t!]{p{2.3cm} p{1cm} p{3cm} p{8cm} p{0.7cm}} 
		\Xhline{3\arrayrulewidth} \vspace{0.1mm}
		\textbf{AIM/ Body part}   & \vspace{0.1mm} \textbf{Input}     & \vspace{0.1mm} \textbf{Methodology}   & \vspace{0.1mm} \textbf{Results} & \vspace{0.1mm} \textbf{Ref}\\  \hline
		Hand Exoskeleton & EEG & Covariate Shift Detection (CSD) based adaptive classifier & The binary classification accuracy was high (p $<$ 0.01). The two-stage CSD test and reported classifier adaptation time was in the order of milliseconds and suitable for real-time application  & \cite{8239668}		\\  \hline

		Ankle       & EMG       & 
		Finding walking style deviation from optimum/ adjusting EMG features
		& 77.2\% four class prediction accuracy. The LDA algorithm used time-domain features for computation efficiency and real-time performance.    & \cite{Gregory2019} \\ \hline
		
		Lower Limb & IMU & Kinematic Iterative Estimation & The algorithm improved assistive force profile and showed $\sim 0.11-0.18N.m$ joint moment deviation. Real-time performance have been achieved by implementing smooth transition assumption (STA) to predict 2D ground-foot contact wrench. & \cite{9606192}		\\ \hline
				
		Gesture recognition  & EMG       & CNN on high-density sEMG & 		89.3, 99.0 and 99.5\% accuracy for eight class classification. These accuracies belong to 1, 40 and 150 frames, respectively, each frame corresponding to 1 ms signal observation time, determining the real-time performance.    & \cite{Geng2016} \\ \hline
		Motion intention  & EEG       & LDA & The movement intention (binary classification) was detected 460 ± 85 ms before actual onset with 92\% accuracy for the paretic arm & \cite{Lew2012} \\ \hline
		Brain-machine interfaces & ECoG, Eye tracking & Hybrid Augmented Reality Multimodal Operation Neural Integration & Up to 92.9\% Complex taskmovement detection accuracy. The median latency between the prediction and the actual movement onset was 200 ms & \cite{McMullen2014} \\ \hline
		Grip prediction & FMG & LDA & Between 70 to 89\% 11-class classification accuracy. A commercial prosthesis (Bebionic 3) has been used for real-time control of the robot using FMG.     & \cite{Cho2016} 
		\\ \hline
		Hand Motions prediction & Ultrasound imaging & Nearest neighbor classification & 91-92\% classification accuracy for 15 hand motions. The total prediction time consist of 200 ms data acquisition time plus $79\pm5$ ms processing time.   & \cite{Akhlaghi2016} \\ \hline
		
		Arm Prosthesis control & EMG and gaze & Two SVRs for x and y position & The algorithm was able to estimate the target position in 2D space with 6.94 cm RMSE. On average 535 ms delay from the beginning of gaze fixation and the end of reaching motion has been reported. This number was 121 ms from end of gaze fixation. &  \cite{Krausz2020} \\ \hline
		
		Brain Computer Interface & EEG & LDA & The proposed method was successfully implemented on \emph{ASSISTON-MOBILE} exoskeleton with 63-84\% binary classification accuracy. The robot is able to decode the subject's intention and robot's state every 300 ms.  & \cite{Sarac2013} \\ \hline
		
		Intention Estimation & Force, position, velocity & NN & In simulations, the desired trajectory was predicted with $\sim 10e-8$ MSE. However, there was always time delay between the robot arm and the desired trajectory caused by phase difference in impedance function.   & \cite{Ge2011} \\ \hline
		
		Wrist Muscles & EMG & Deep Belief Networks & 88.6\% accuracy for five class classification. The 166ms signal observation time helps with real-time performance as long as this number plus the computation time is less than 300 ms \cite{Englehart2003}.  & \cite{Shim2015} \\ \hline
		
		Multimodal data & EEG-EMG & Deep autoencoder  & They achieved up to 78.1\% accuracy in predicting joint compression (binary). Computation cost has not been reported.  & \cite{Said2017} \\ \hline
		
		Lower limb & EMG & Deep Belief Networks & They managed to reduce the joint angle RMSE by 50\% by using DBN on multimodal sEMG compared to PCA method. Computation cost has not been reported   & \cite{Chen2018} \\ \hline
		
		Shoulder and arm & EMG & Recurrent CNN & Estimate 3D motion trajectory with 91.7\% $R^2$ value. The CNN model uses shorter signal observation time, 150 ms compared to 200 ms for support vector regression (SVR) models which is better for real-time applications. However it is computationally more expensive. & \cite{Xia2018} \\ \hline
		
		Shoulder and Elbow & EMG & Time Delayed Neural Networks &  The model was able to estimate joint torque with 2.17 and 1.19 N.m RMSE for shoulder and elbow, respectively. The 200 ms window size helps with real-time performance, however, the computation time was not reported.   & \cite{Loconsole2014} \\ \hline
		
		Upper limb & EMG & TD features/ PCA/ MLP & Estimates force with 0.9 $R^2$ performance in 100 ms signal epoch.   & \cite{Nielsen2011} \\ \hline
		
		Knee & Inertial \& fiber optic 
		& Kalman filter of two IMUs and variation-based POF curvature & 
		The RMSE between the POF-IMU sensor system and the encoder was between 1 to 4\textdegree. The computation cost has not been reported.
		& \cite{Leal-Junior2018} \\ \hline
		
		Lower Limb Prosthesis & prosthesis sensors & Time-based features extraction, PCA, Gaussian Mixture Models (GMM) & Predicted 90 activities with 100\% accuracy without perceivable delay by the user. The intent recognition delay was 500 ms at 100 ms window size, however, as mid-level controllers work based on system dynamics and mechanical sensors, this delay did not hurt real-time performance.     &   \cite{Varol2010} \\ \hline
		
		
		Spinal Exoskeletons & IMU & Gaussian mixture and state machine & Q-Passive assist with 86.7\% accuracy. The highest accuracy for GMM model has been observed at 0.3 normalised time. Computation time has not been reported.   & \cite{Jamsek2020} \\ \hline
		
		Gait phase estimation & Hip angles & Adaptive Oscillator & 97-99\% accuracy, $PE \sim -21.12$, $MSE \sim 0.05$, and $Lead Time \sim -75 ms$   & \cite{8794856}		
		
		\\ \Xhline{3\arrayrulewidth}
	\end{tabular}
\end{table*}

\subsubsection{Signal processing}
Sensor signals carry noises from different sources. For instance, EMG signals have electrode inherent noise, movement artefact, electromagnetic noise \cite{Reaz2006}, muscle cross-talk \cite{Gerdle1999} internal body noise and electrocardiographic (ECG) artifacts for upper-chest recording \cite{Willigenburg2012}. Different strategies can be implemented to minimise the effects of these noises. For instance, increasing the electrode size will reduce the impedance and give a high SNR. The movement artefact and electromagnetic noise could be minimised using a high and low pass filter.

After filtering noises, the signal needs to be processed before feeding into the intent prediction model. The raw signals are usually large \cite{Triwiyanto2020}, so using them directly for classification can reduce the efficiency of the classifier \cite{Chowdhury2013}. Extracting features from the raw signal can reduce dimensionality while maintaining the most important information. This improves performance and allows real-time computing. Feature extraction can be done in time-, frequency-, or time-frequency domains \cite{AsghariOskoei2007,Phinyomark2012a}. The advantage of time-domain features is that they are easier to calculate with less computational effort \cite{El-Badawy2022}; hence they are popular in engineering and medical practices \cite{Phinyomark2012a}. To process signals in the frequency domain, Fast Fourier Transform (FFT) helps to shift from the time domain into the frequency domain. The common frequency-domain characteristics include mean frequency (MNF), mean power frequency (MPF) and median frequency (MDF).

\subsubsection{Data fusion}
Data fusion is the combination of information from multiple sources to increase the intent detection performance and accuracy \cite{White1987}. This information can be either redundant to improve reliability or complementary to decrease uncertainty. Data fusion could happen in measurement, features extraction or classification and decision making \cite{Castanedo2013}.

The intent can be acquired through multiple sensors of the same or different modalities. An example of the first one is the knee exoskeleton developed by Lyu et al., which uses 16 EMG sensors on the thigh to predict the knee movement's intention \cite{Lyu2019}. The latter has been implemented by Krausz et al. in which they fused EMG signals with gaze data to control prosthetic arms \cite{Krausz2020}. Researchers at the University of Espírito Santo combined inertial measurement unit (IMU) and intensity variation polymer optical fibre to develop a highly reliable angle measurement system \cite{Leal-Junior2018}. Young et al. studied the effect of combining EMG signals with position, load, and inertial sensors to increase pattern recognition accuracy\cite{Young2014}.

\subsection{Intent interpretation}
For many scenarios, the direct measurement of the intent is impossible; however, we can measure the related parameters \cite{Losey2018}. The interpretation, then, relies on the modelling and simplification of assumptions \cite{Schmiedeler2018}. Either pattern recognition or continuous effort mapping can help with this mapping \cite{Losey2018}. Pattern recognition uses AI algorithms to map the input signals or their reduced features to a set of functions. Some of the widely used algorithms in the field are linear discriminant analysis (LDA) \cite{Gregory2019}, support vector machine (SVM) \cite{Yin2021}, artificial neural networks (ANN) \cite{Jali2016} and fuzzy neural networks (FNN) \cite{Kiguchi2007}. The general approach here is feature engineering or machine learning (ML), in which the data shall be pre-processed and standardised. Then the features should be extracted and reduced in dimensionality and fed into the model to do the classification \cite{Wang2020a}. Fig. \ref{dm} shows the intent interpretation schematic of a wearable assistive device.

With the increasing number of available data as big data in recent years, deep learning (DL) has become popular in the field. Compared to general ML methods like shallow learning, DL can extract high-level features from low-level inputs \cite{Phinyomark2018}. For instance, Convolutional Neural Networks (CNN) perform exceptionally good at finding local features. Geng et al. used CNN on sEMG data to recognise finger motions and achieved up to 99.5\% accuracy from 150 ms signal observation time \cite{Geng2016}. Recurrent Neural Networks (RNN) use feedback connected to previous layers to store prior inputs to model the problem in time. The most popular RNN models are long short-term memory units (LSTM) and gated recurrent units (GRUs). To take advantage of both RNN and CNN, a combination model (RNN + CNN) has been proposed \cite{Xia2018}. The RCNN model uses a shorter signal observation time, 150 ms, compared to 200 ms for support vector regression (SVR) models, which is better for real-time applications. However, it is computationally more expensive, adding to computation time. Laezza showed that RNN performs the best myoelectric signal classification among RNN, CNN and RNN+CNN \cite{Laezza2018}.

\begin{table*}
	\centering
	\footnotesize
	\caption{The most renowned passive exoskeletons in the literature} \label{tbl:pss}
	\begin{tabular}[tbh!]{p{5cm} p{2cm} p{2cm} p{5cm}}
		\hline \vspace{1mm}
		
		Passive Laevo \cite{Bosch2016}:
		
		- Forward-bended postures.
		
		- Reduces muscle activity and discomfort.
		
		- Increases endurance time.
		
		- Uses an elastic tube to transfer the force from the lower back to the chest and legs.
		& \centering
		\raisebox{-\totalheight}{\includegraphics[scale=0.13]{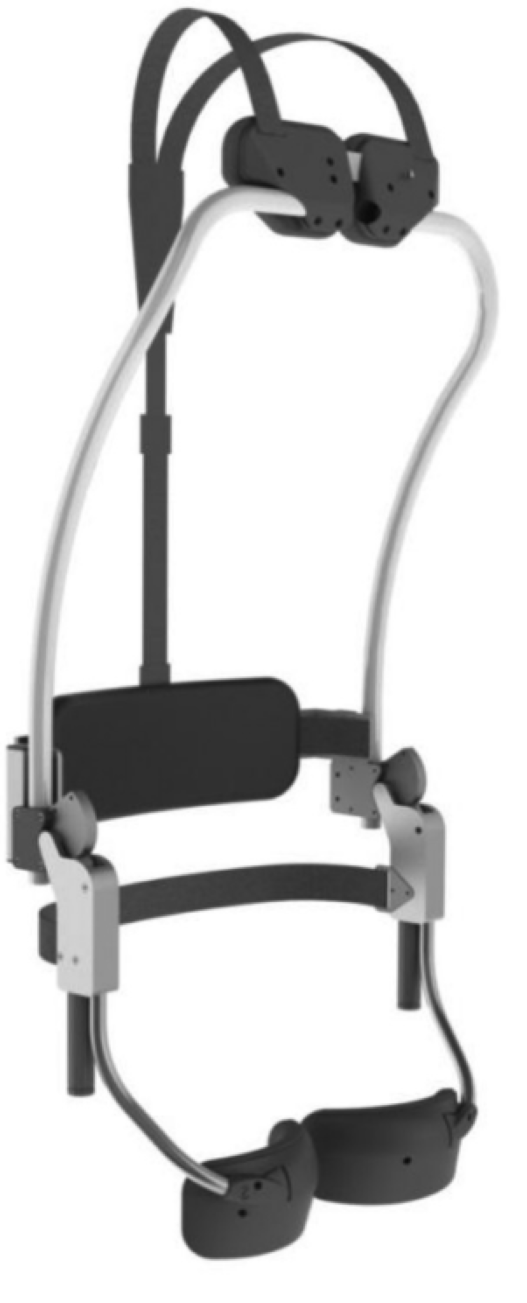}}
		& \centering
		\raisebox{-\totalheight}{\includegraphics[scale=0.2]{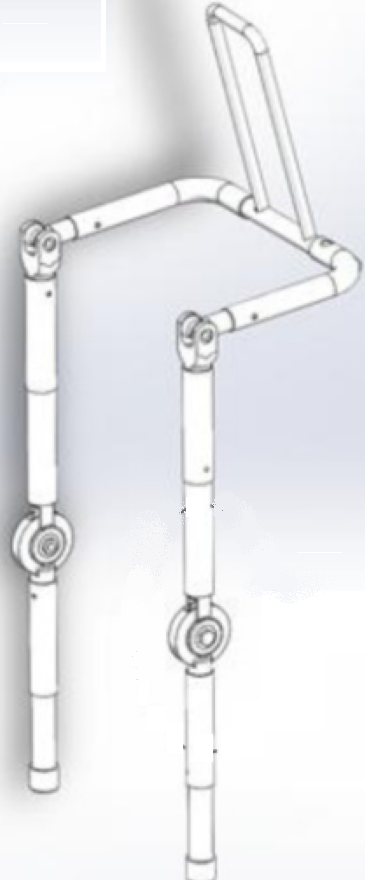}}
		& \vspace{1mm}
		
		Passive Knee-Assisting Exoskeleton (PKAExo)\cite{Li2018}:
		
		- Reduces the maximum knee moment.  
		
		- A wirerope connected to a press spring, going around a eccentric pulley at the knee placement. 
		\\  \vspace{3mm}
		
		Single leaf exoskeleton \cite{Grabowski2009}:
		
		- Reduces the hopping metabolic rate.
		
		- Uses set of fiberglass springs parallel with the user's legs.
		& \centering
		\raisebox{-\totalheight}{\includegraphics[scale=0.1]{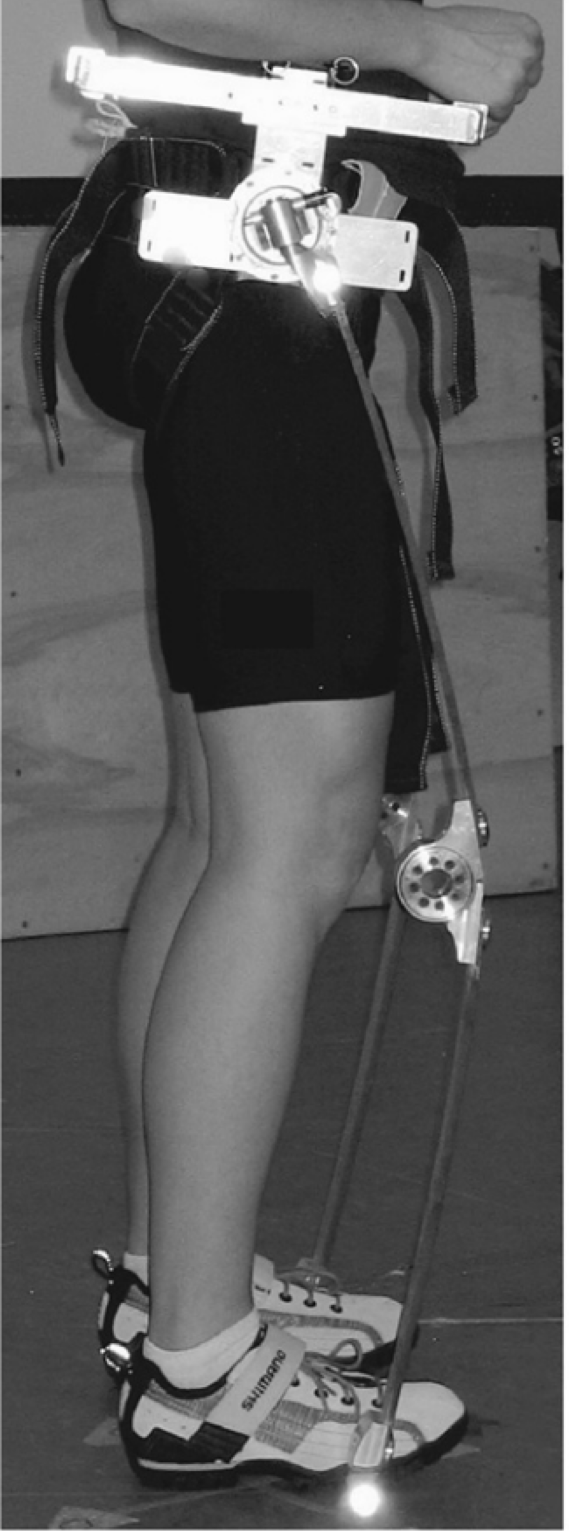}}
		& \centering 
		\raisebox{-\totalheight}{\includegraphics[scale=0.15]{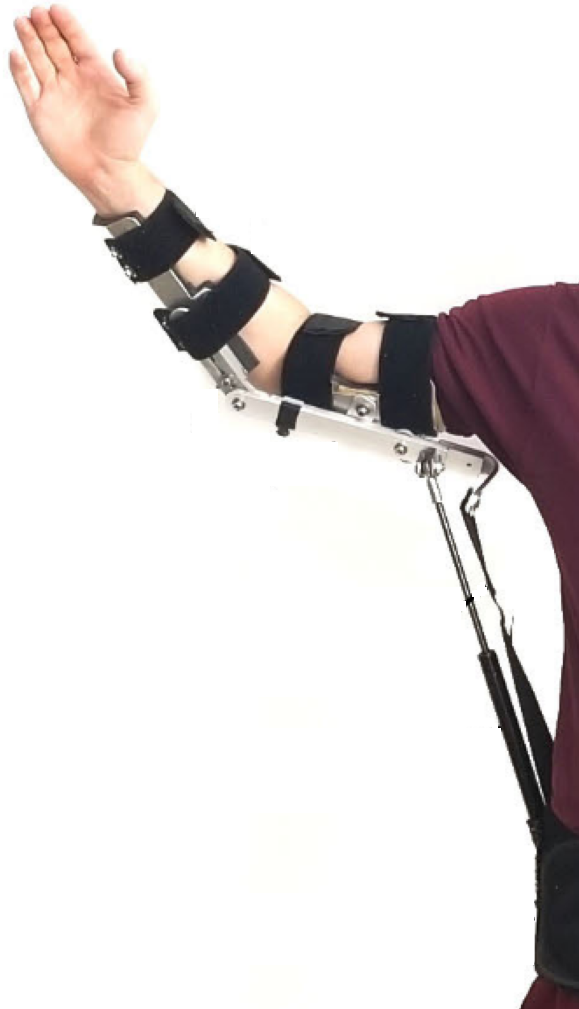}}  
		& \vspace{4mm}
		
		Panto-Arm Exo \cite{Hull2020}:
		
		- Gravity compensation arm support exoskeleton. 
		
		- Reduce the activity of the Biceps Brachii by up to 52\%. 
		\\ 
		
		Unpowered exoskeleton \cite{Collins2015}
		
		- Uses a spring to connect foot and shank.
		
		- A passive clutch, hinged at the ankle 
		
		- Helps with the calf muscles and Achilles tendon.
		& \centering
		\raisebox{-\totalheight}{\includegraphics[scale=0.14]{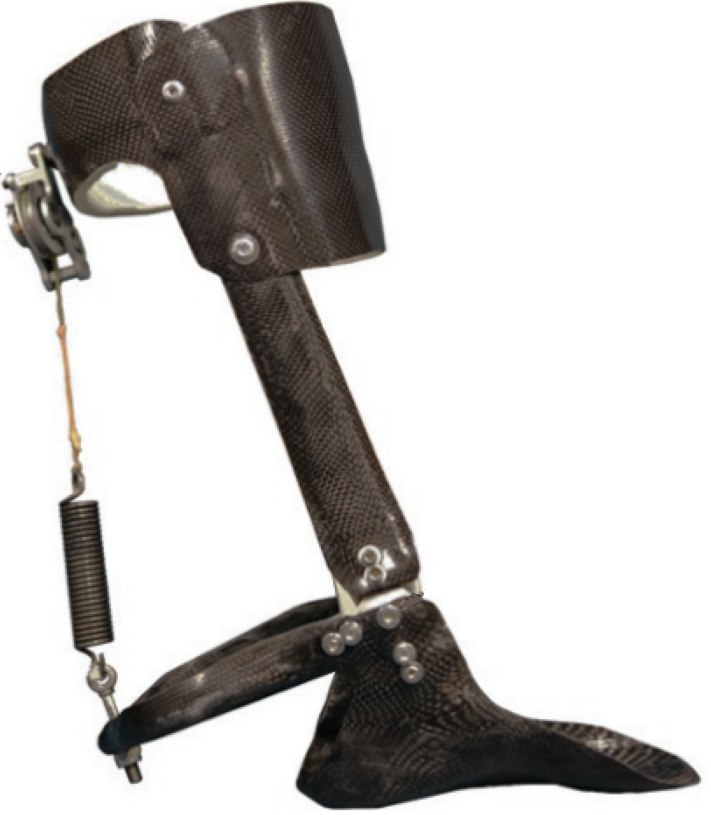}}  
		& \centering
		\raisebox{-\totalheight}{\includegraphics[scale=0.15]{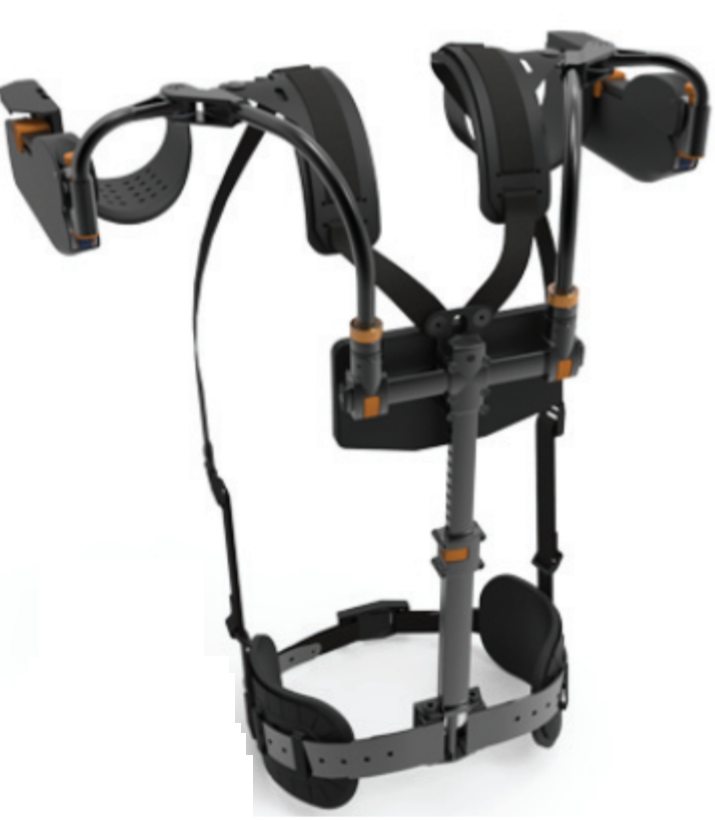}}  
		& \vspace{1mm}
		
		Levitate \cite{SPADA20171255}:
		
		- Supports workers' arms.
		
		- Transfers the weight of the upper limb to the metallic core.
		\\ \vspace{1mm}
		
		Occupational exoskeleton \cite{Moyon2019}:
		
		- Helps with arm-elevated tasks.
		
		- Uses two flat springs in the back.
		
		- Reduces upper body strain.
		& \centering
		\raisebox{-\totalheight}{\includegraphics[scale=0.16]{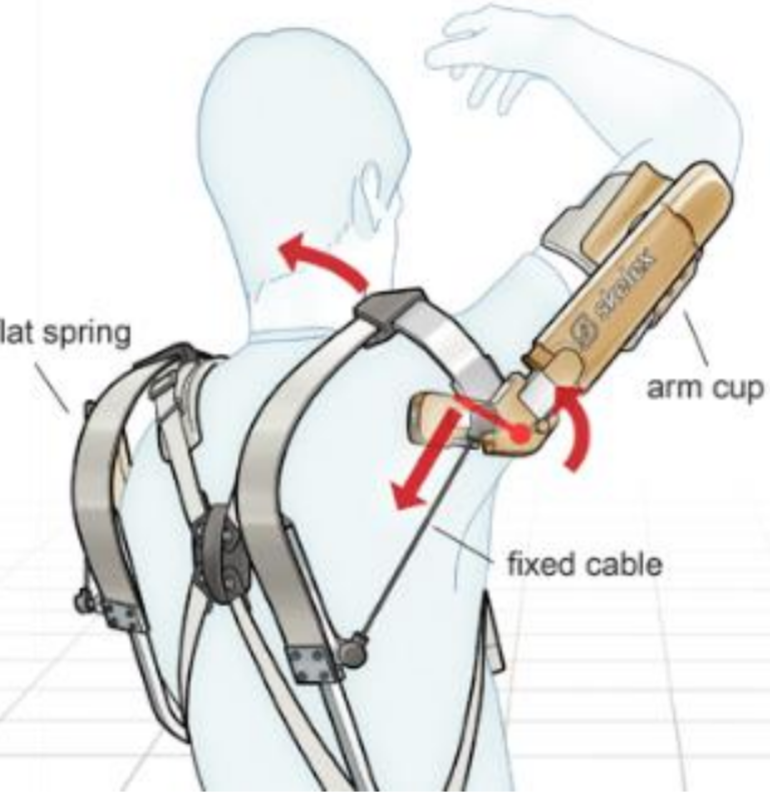}}  
		& \centering
		\raisebox{-\totalheight}{\includegraphics[scale=0.15]{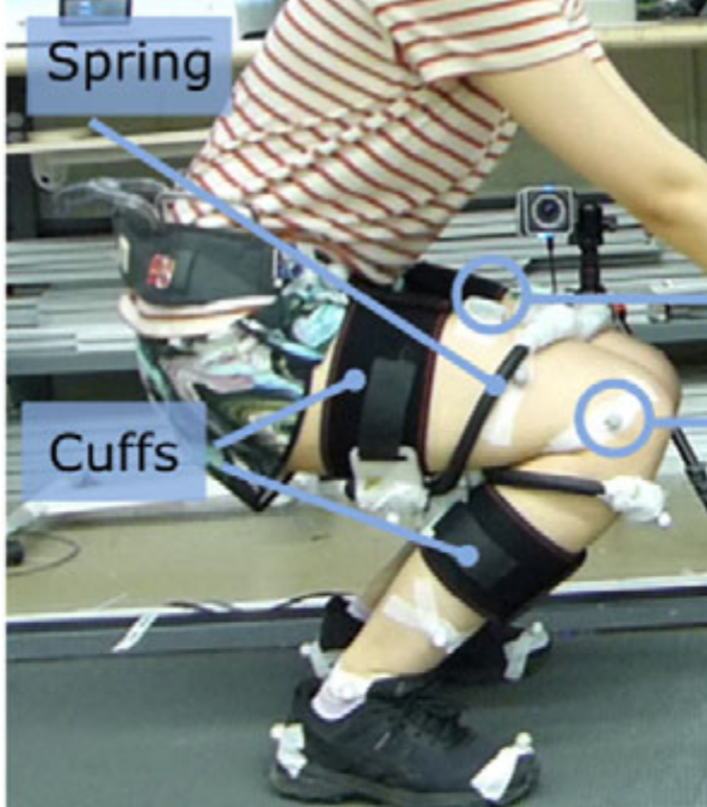}}  
		& \vspace{1mm}
		
		SpringExo \cite{Hidayah2021}:
		
		- Helps with knee extension during squat.
		
		- A spring coil connects thigh and shank cuffs to store the knee flexion energy.
		\\ \hline
	\end{tabular}
\end{table*}

\looseness=-1 Unsupervised pre-trained networks (UPN) are another category of DL models, which can be stacked auto-encoders or deep belief networks (DBN). The first one comprises multiple auto-encoder neural network layers to find the hierarchical features. The DBN uses restricted Boltzmann machine layers to find the joint probability distribution of the training data. Shim and Lee achieved 88.6\% classification performance using DBN, which was 2.9 and 7.6\% higher than the support vector machine (SVM) and linear discriminant analysis (LDA) \cite{Shim2015}. UPNs have also shown the potential to replace traditional unsupervised feature projection methods. Said et al. developed a DL model for EMG and EEG signal classification, showing that multimodal autoencoder has less distortion and outperforms unimodal algorithms in classification accuracy \cite{Said2017}. Chen et al. used DBN to estimate joint angles of the hip, knee and ankle and reached a 50\% reduction in RMSE compared to the principal components analysis (PCA) method \cite{Chen2018}.

Continuous effort mapping can be helpful when extracting a discrete set of functions does not satisfy the control system's requirements, and continuous time-varying intention is necessary. An example of this could be determining the joint torque based on the amplitude of the EMG signal of a muscle connected to that joint \cite{Song2008}. For this type of control, human adaptability can help minimize the error by tuning the EMG activation, proving that proportional control is sufficient for observed robot motion \cite{Lenzi2012}. 

Continuous effort mapping can be categorised into model-based \cite{Peternel2016} and model-free \cite{Vujaklija2018} approaches. Model-based approaches help to map the inputs to the desired outputs using dynamic, kinematic or musculoskeletal models that are precisely developed around human biomechanics. Koike et al. used ANN to build a dynamic model of arm movement using EMG signals \cite{Koike1995}. Wang and Buchanan were the first ones to use the Hill-type model to calculate muscle forces \cite{Wang2002}. On the other hand, model-free approaches use machine learning techniques to learn the relationship between the input and the desired output, known as the "black box" or "unknown". Some popular methods are ANN, fuzzy approximation, Bayesian network, hidden Markov model, and Kalman filter. Loconsole et al. utilised a time-delayed neural network for an exoskeleton to estimate the shoulder and elbow torque of the robot from the mean absolute value of the EMG signal \cite{Loconsole2014}. Nielsen et al. estimated contralateral limb force by multilayer perceptron ANN from four features of 100ms EMG signal episodes to control multi-DoF prostheses \cite{Nielsen2011}. Table \ref{tbl:hmi} highlights some recent efforts on intent prediction.

\section{TYPES OF EXOSKELETONS}~\label{sec:types}
Exoskeletons can be categorised based on their application, structural design or power consumption. The most common classification is based on passive, semi-passive or active mechanisms.

\subsection{Passive exoskeletons }
Passive exoskeletons work by saving muscle energy in a spring or other elastic element and releasing it when needed. This type of exoskeleton does not need actuator and has fewer moving parts than active or semi-passive ones, making them a less complicated, lightweight, and economical option \cite{Li2018}. One of the first models of these passive wearable assistive devices is Yagn's walking aid which utilises two leaf springs parallel to the legs \cite{Yagn1890}. Passive exoskeletons can significantly reduce muscle load and metabolic cost if they are designed optimally \cite{VandenBogert2003}.

\begin{table*}
	\centering
	\footnotesize
	\caption{Semi-passive devices in the literature} \label{tbl:semi}
	\begin{tabular}[tbh!]{p{5.5cm} p{2.1cm} p{2.1cm} p{5.5cm}}
		\hline \vspace{1mm}
		
		H-PULSE \cite{Grazi2020a}:
		
		- Upper-limb work assist exoskeleton.
		
		- Adjusts assist ratio based on shoulder joint angle and user selected level.
		& \centering
		\raisebox{-\totalheight}{\includegraphics[scale=0.19]{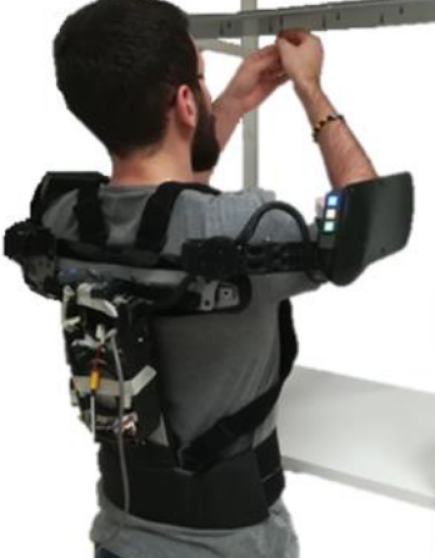}}  
		& \centering
		\raisebox{-\totalheight}{\includegraphics[scale=0.23]{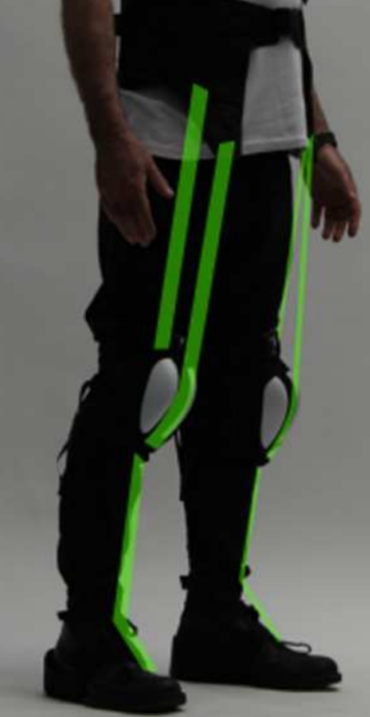}}  
		& \vspace{1mm}
		
		Garment by XoSoft \cite{10.3389/fnbot.2020.00031}:
		
		- Uses elastic bands store and release energy.
		
		- Helps patients with mobility impairments. 
		
		- Uses foot contact and knee angle to identify the gait phase.
		\\ 
		
		Quasi-passive ankle exoskeleton \cite{Kumar2020}:
		
		- Uses model-free, discrete-time extremum seeking control to optimise the spring's stiffness.
		
		- Stores early and mid-stance energy  and releases it at ankle push-off.
		& \centering
		\raisebox{-\totalheight}{\includegraphics[scale=0.12]{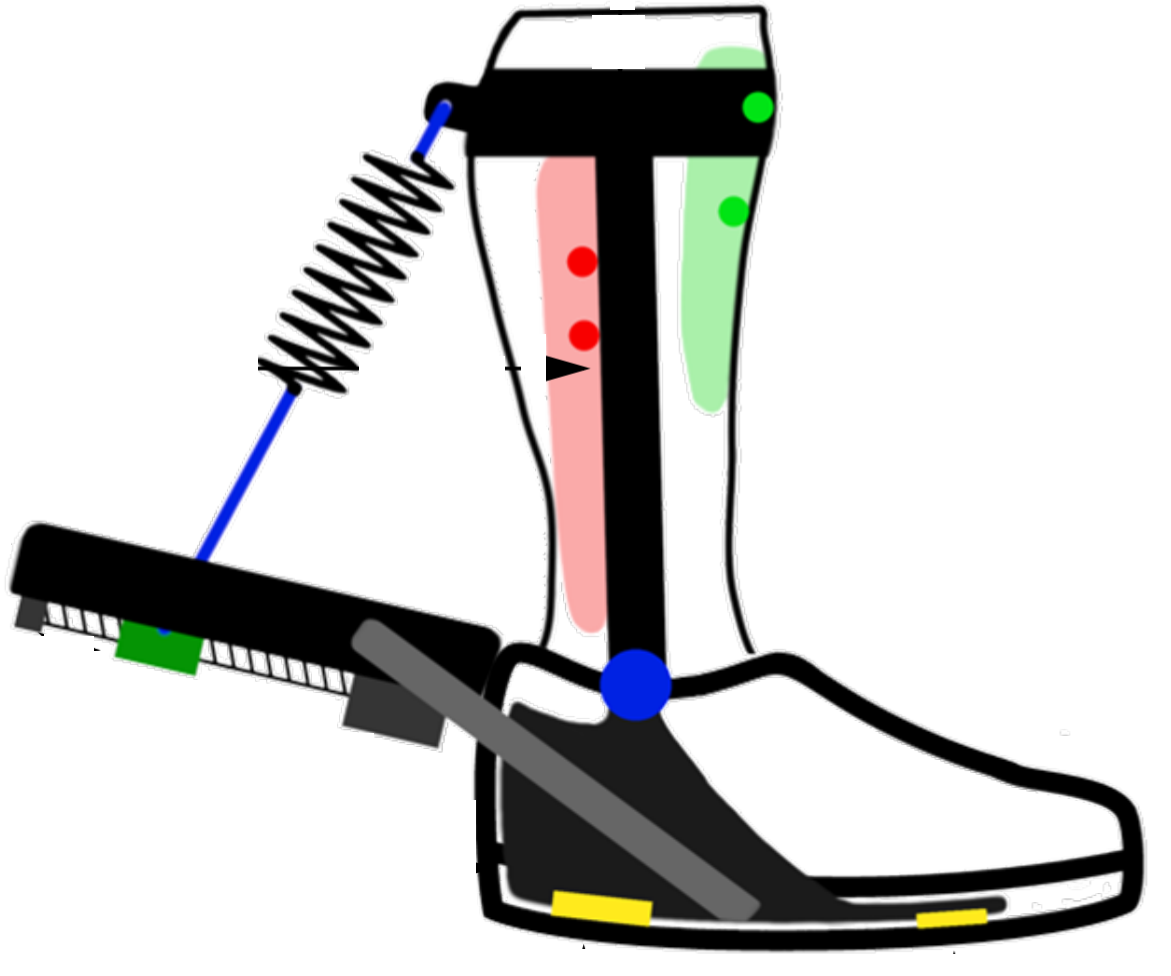}}  
		& \centering
		\raisebox{-\totalheight}{\includegraphics[scale=0.16]{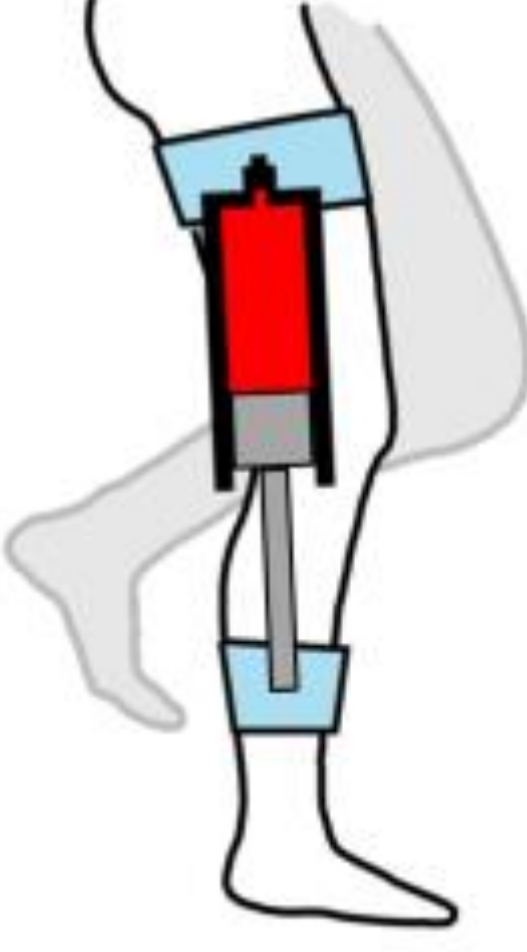}}  
		& \vspace{1mm}
		
		Quasi-passive knee exoskeleton \cite{Rogers2017}:
		
		- Helps with knee extensors during decent.
		
		- Uses air spring.
		
		- Control solenoid valve and seals on the heel strike
		\\ 
		
		Quasi-passive lower limb exoskeleton \cite{Chang2020}:
		
		- Transfers knee work in late swing phase to help with ankle push-off in mid-stance phase. 
		
		- Uses two servo-controlled clutch at each end of the spring for timing.
		& \centering
		\raisebox{-\totalheight}{\includegraphics[scale=0.22]{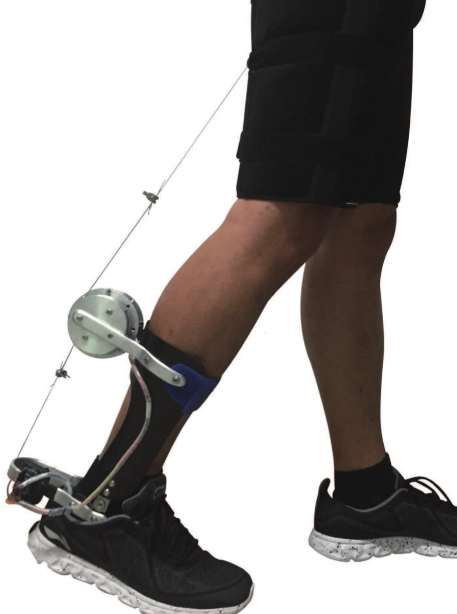}}  
		&  \centering
		\raisebox{-\totalheight}{\includegraphics[scale=0.21]{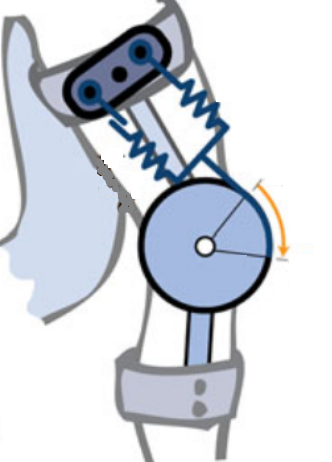}}  
		& 
		
		Quasi-passive knee exoskeleton \cite{Shamaei2014}:
		
		- Identifies the gait phase using contact switches at the heel and toe.
		
		- Finite-state machine controls a friction-based latching mechanism to dis/engage a spring and achieve two levels of stiffness.
		\\  \vspace{1mm}
		
		Quasi-passive artificial gastrocnemius \cite{Eilenberg2018}:
		
		- A clutched-spring system paired with a powered ankle prosthesis.
		
		- Helps with knee and ankle.
		& \centering
		\raisebox{-\totalheight}{\includegraphics[scale=0.16]{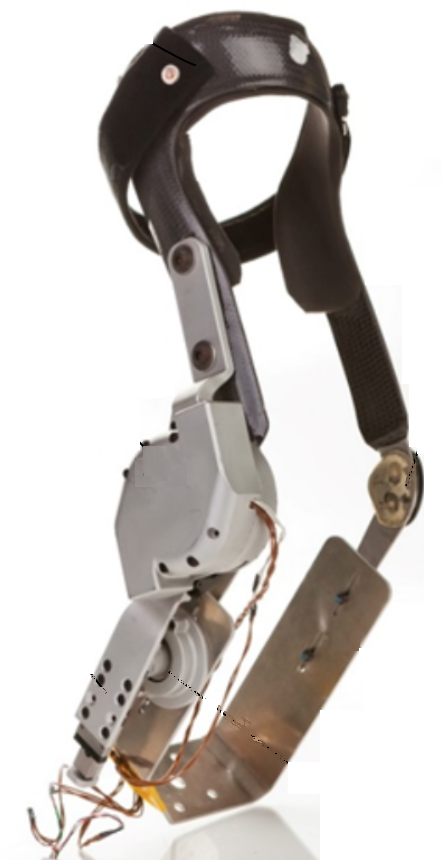}}  
		& \centering
		\raisebox{-\totalheight}{\includegraphics[scale=0.16]{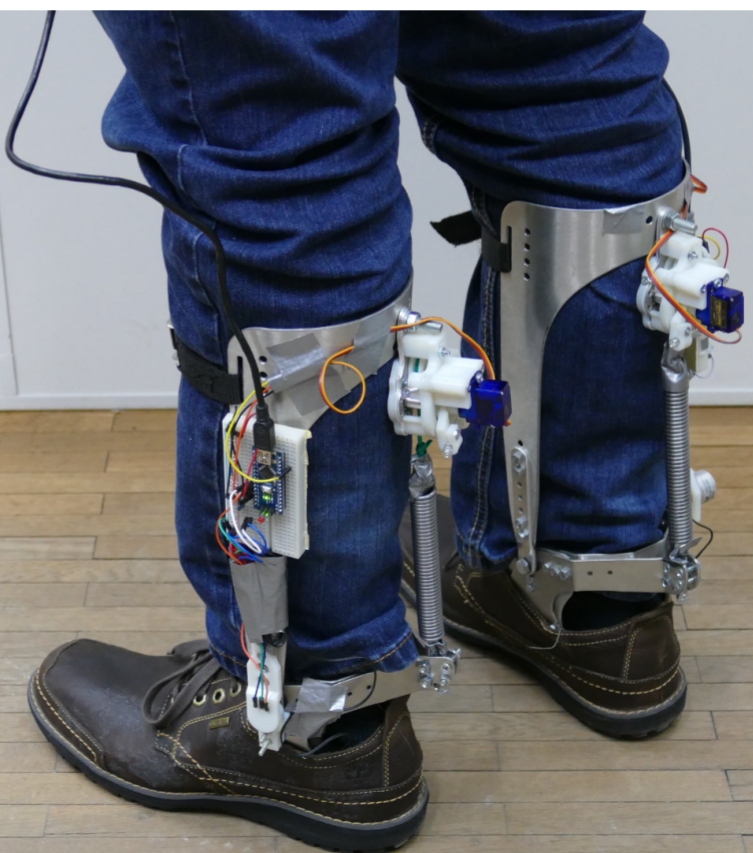}}  
		& \vspace{1mm}
		
		Clutch-Actuated Ankle Exoskeleton \cite{Dezman2018}: 
		
		- Uses a servo controlled clutch connecting to a linear spring. 
		
		- Arduino controls the clutch based on heel strike.
		\\ \hline
	\end{tabular}
\end{table*}


There are different approaches to designing a passive exoskeleton. A common way is to transfer a rotating joint's kinetic energy into an elastic element. Collins et al. developed a passive ankle assistive device that helps the wearer with calf muscle activity \cite{Collins2015}. It uses a spring to connect the shank to the calf for energy storage. A clutch engages the base on the ground and disengages when it is in the air. The results show a 7.2 ± 2.5\% reduction in healthy human metabolic walking cost. Another example is the passive knee assist exoskeleton (PKAExo) for weight climbing \cite{Li2018}. Its eccentric pulley design allowed for a nonlinear increase in the assist ratio with knee angle. It showed a decrease of 21\% in the average number of the knee extensor's maximum load.

Another approach to passive exoskeletons lies in exotendons. In some animals, muscles connect multiple joints \cite{Biewener1998} which helps in energy saving when one joint requires acceleration and the other needs deceleration \cite{VanIngenSchenau1990}. Needless to say that the elasticity of the muscles and tendons helps with the storage and release of energy \cite{Alexander1977,Dickinson2000}. This can help animals to use only 50\% metabolic energy compared to humans \cite{Biewener1998}. Single leaf exoskeleton (SLE) from MIT utilises this concept and demonstrates a reduction of 24\% in metabolic cost during hopping \cite{Grabowski2009}.

Another solution is using pulleys and more complicated setups. Bogert developed a passive exoskeleton concept with elastic cords and pulleys that spans multiple joints \cite{VandenBogert2003}. The idea was to assist with walking by transferring the saved energy into different joints. In simulations, in the most straightforward setup with only uniarticular exotendons at the ankle, a muscle torque reduction of 21\% was achieved. By replacing the exotendons with triarticular ones, spanning the ankle, knee and hip, this reduction increased to 46\%. It has been shown that a maximum of 71\% muscle torque reduction could be achieved by using multi exotendons in complex configurations. Based on this concept, XPED2 was designed to assist with lower-limb \cite{VanDijk2014}. In this system, a cable connects a leaf spring at the foot to the pelvis through a pulley at the knee. Change in the hip and ankle joint angle results in the spring's deformation, thus force in the cable. The experimental result showed 12.1\% assistance and a slight increase in the metabolic cost. Dijk and Kooji concluded that the 71\% assistance in Bogert's is probably unlikely to achieve \cite{VanDijk2014}. Table \ref{tbl:pss} shows some of the most renowned passive exoskeletons in the literature.

\begin{table*}
	\centering
	\footnotesize
	\caption{Examples of active assistive devices built in the past.} \label{tbl:active}
	\begin{tabular}[tbh!]{p{5.7cm} p{1.8cm} p{1.8cm} p{5.7cm}}
		\hline \vspace{1mm}
		
		Hybrid Assistive Limb (HAL) \cite{Sankai2010b}:
		
		- Improve user's capabilities
		
		- Uses DC motors 
		
		- Actively controls hip and knee, and spring for ankle. 
		
		- Two complementary control systems based on EMG data and robot and user's condition.
		& \centering \raisebox{-\totalheight}{\includegraphics[scale=0.16]{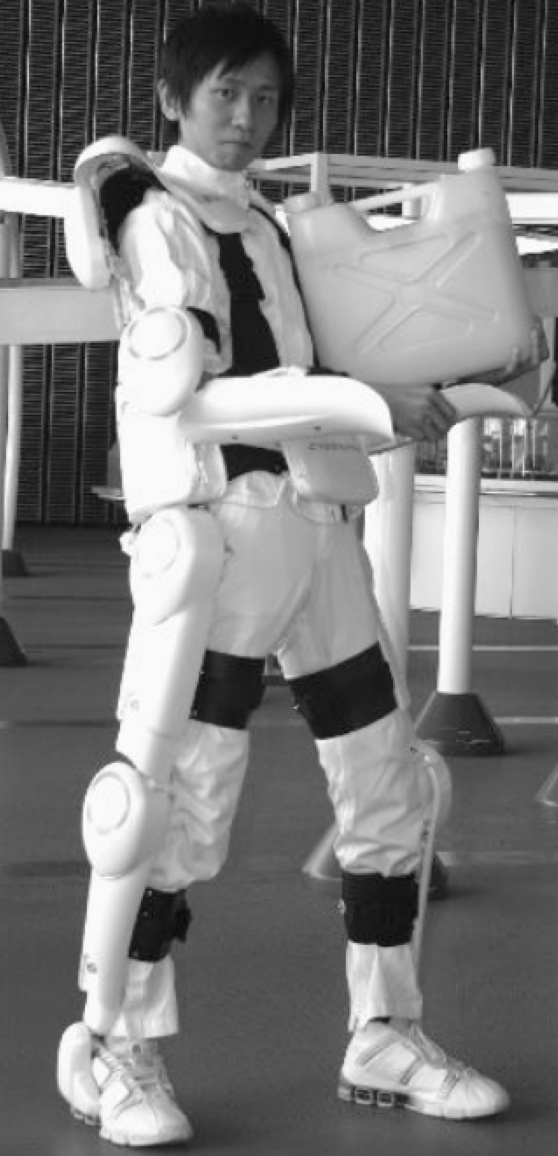}} 
		& \centering
		\raisebox{-\totalheight}{\includegraphics[scale=0.1]{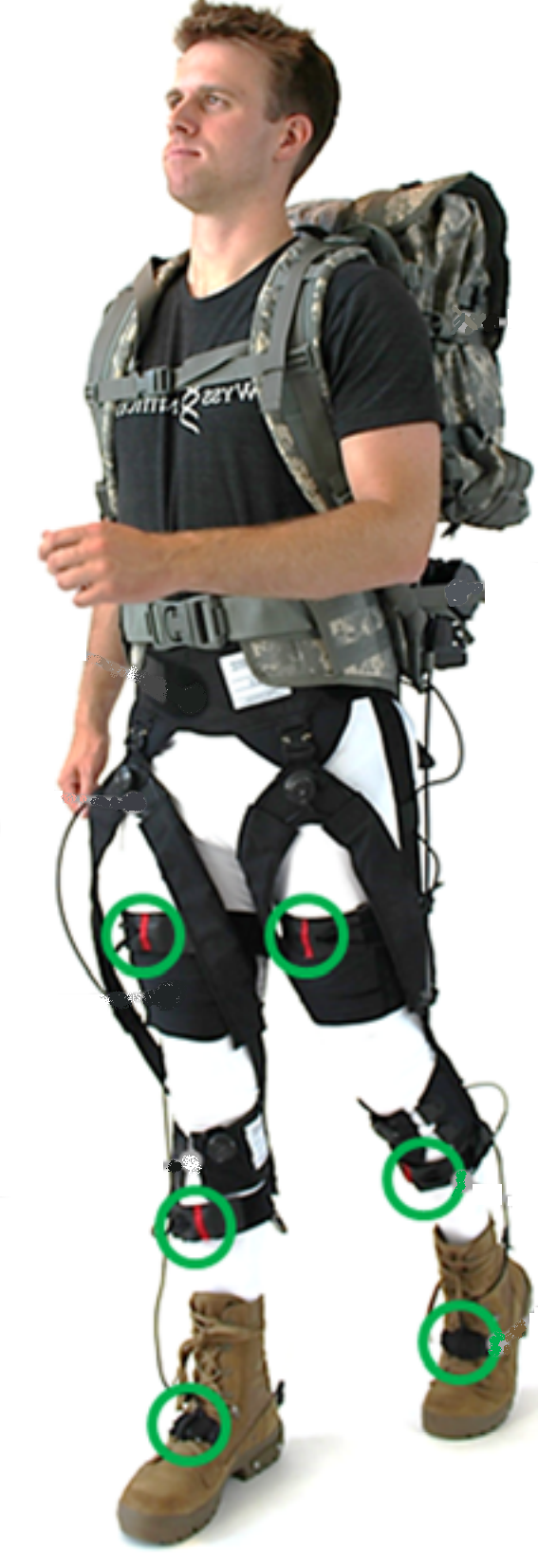}}  
		& \vspace{1mm}
		
		Multi-joint soft exosuit by Wyss Ins. \cite{Lee2018}: 
		
		- Uses iterative force-based position control of the Bowden cables. 
		
		- Multi-articular controller detects the heel strike and assists with ankle push-off. 
		
		- Hip extension controller helps in early stance using constant timing parameters. 
		
		- Uses user's bones for compression forces.
		\\ \vspace{1mm}
		
		MINDWALKER \cite{Gancet2012}: 
		
		- Brain controlled lower limb exoskeleton. 
		
		- Uses EEG control system backed with EMG complementary. 
		
		- Optimised to minimise the walking cost.
		& \centering \raisebox{-\totalheight}{\includegraphics[scale=0.15]{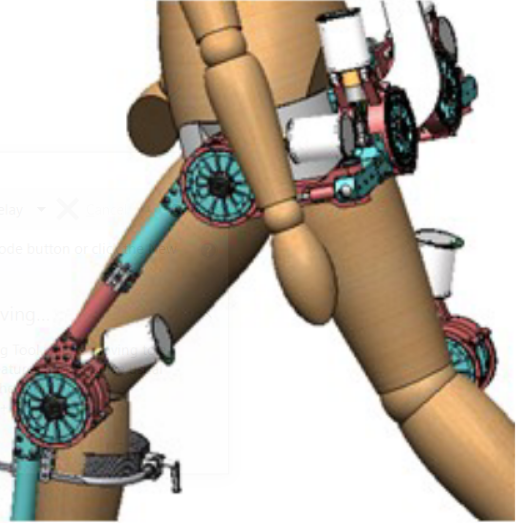}}  
		& \centering \raisebox{-\totalheight}{\includegraphics[scale=0.13]{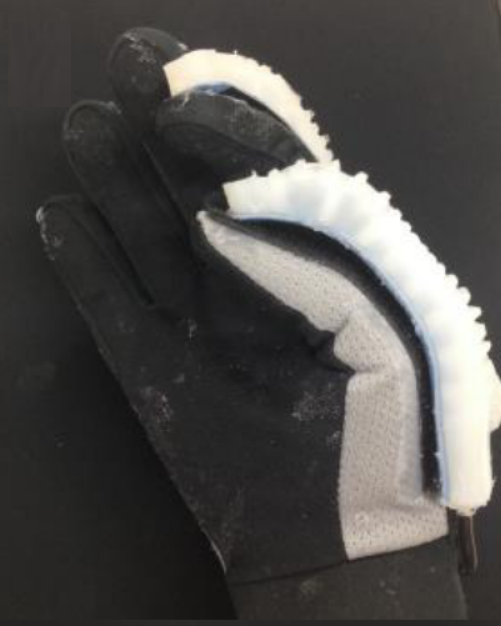}}  
		&  \vspace{1mm}
		Soft pneumatic hand exoskeleton \cite{Yap2015}:
		
		- Hand therapy.
		
		- Uses different pneumatic soft actuators with variable stiffness embedded in a glove  to achieve different bending profiles.
		\\  \vspace{1mm}
		
		ATLAS \cite{Sanz-Merodio2012}:
		
		- The user manually controls general commands. 
		
		- Impedance controller follow normal gait trajectory by measuring the condition.
		& \centering \raisebox{-\totalheight}{\includegraphics[scale=0.17]{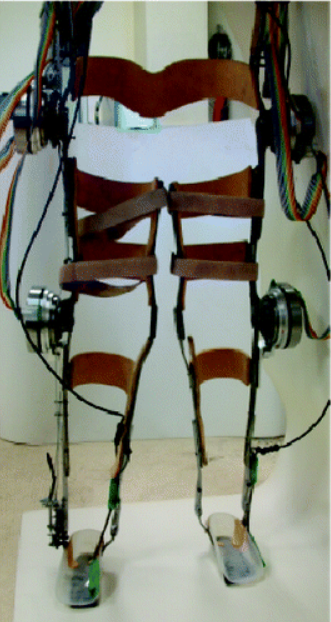}} 
		& \centering \raisebox{-\totalheight}{\includegraphics[scale=0.1]{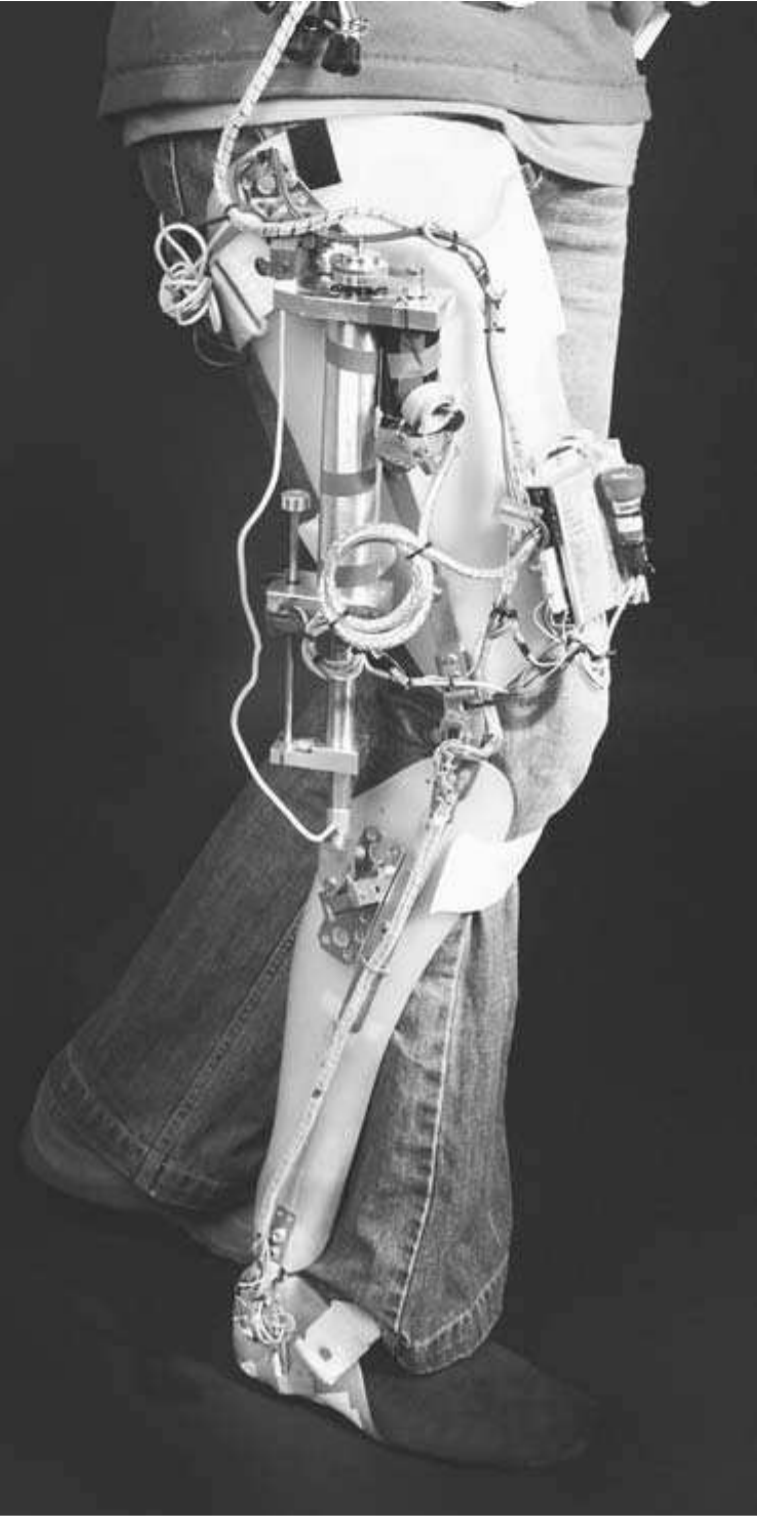}}
		& \vspace{1mm}
		
		Lower-extremity exoskeleton \cite{Fleischer2006}:
		
		-  Motion support in disabled people
		
		- Uses force, Hall and EMG sensors.
		
		- Uses a linear actuator at the knee joint.
		\\ 
		
		Anthropometrically parameterized lower limb exoskeleton \cite{Laubscher2021}:
		
		- Helps with hip and knee in sagittal plane.
		
		- Uses brushless DC motors driven by servo-amplifiers 
		
		- Proportional-integral-derivative controller. 
		& \centering \raisebox{-\totalheight}{\includegraphics[scale=0.15]{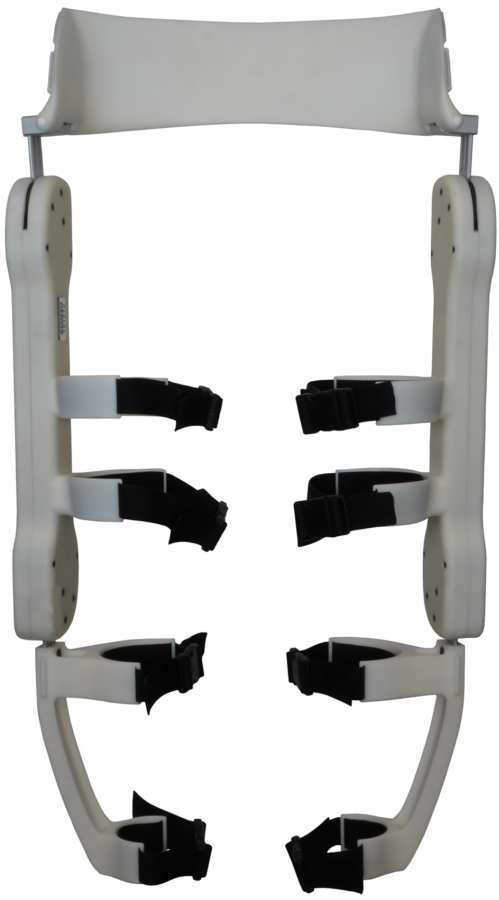}} 
		& \centering \raisebox{-\totalheight}{\includegraphics[scale=0.15]{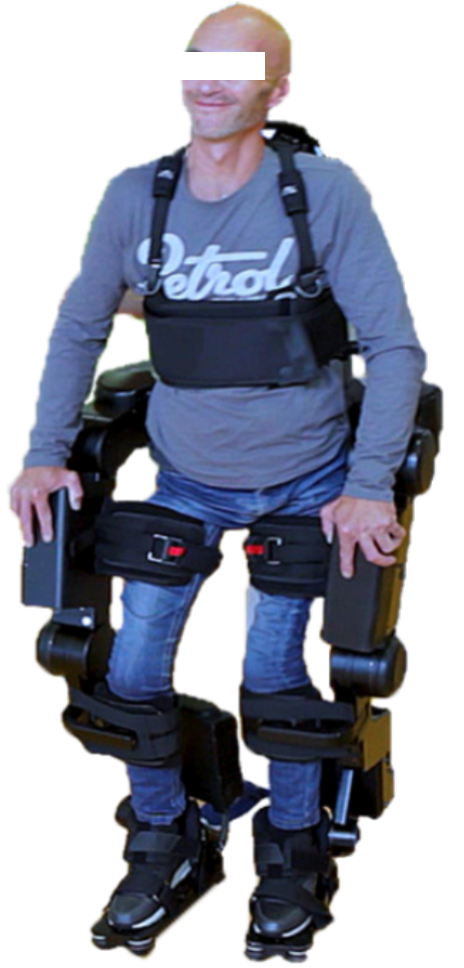}}
		& 
		
		ATALANTE \cite{Paredes2020}:
		
		- 6 active joint in each leg controlled.
		
		- Uses linear actuators and brushless DC motors. 
		
		- Virtual-constraints-based regulation framework on zero moment point.
		\\ \vspace{1mm}
		
		Walking Assist \cite{Font-Llagunes2020}:
		
		- Modular light weight exoskeleton. 
		
		- Helps people with spinal cord injury . 
		
		- IMU and and angular encoder.
		
		- Brushless DC motor.
		& \centering \raisebox{-\totalheight}{\includegraphics[scale=0.08]{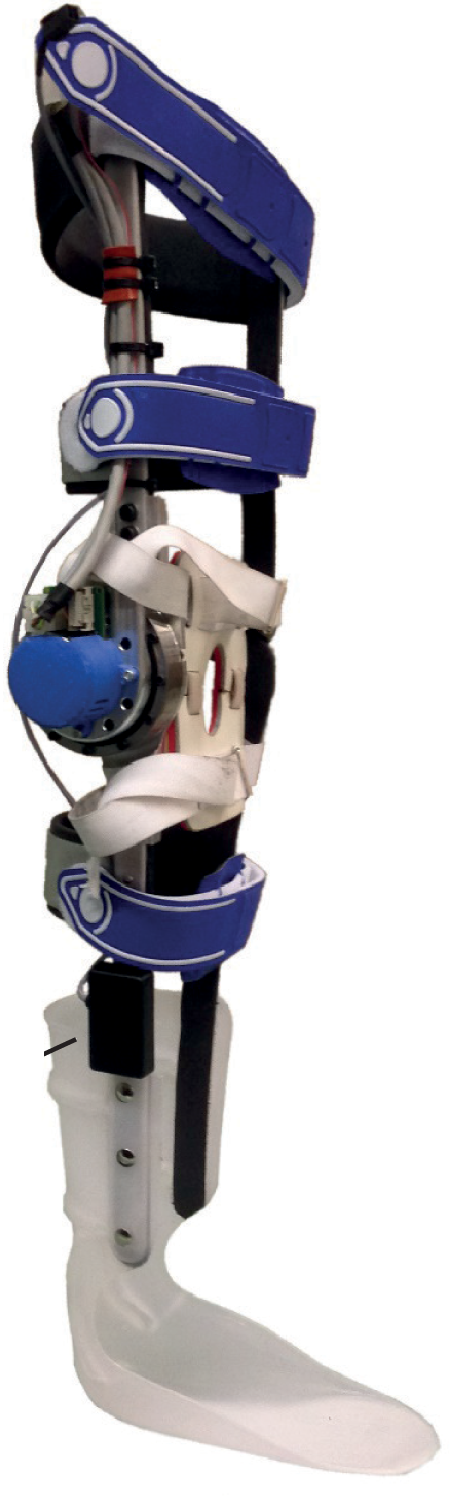}} 
		& \centering
		\raisebox{-\totalheight}{\includegraphics[scale=0.17]{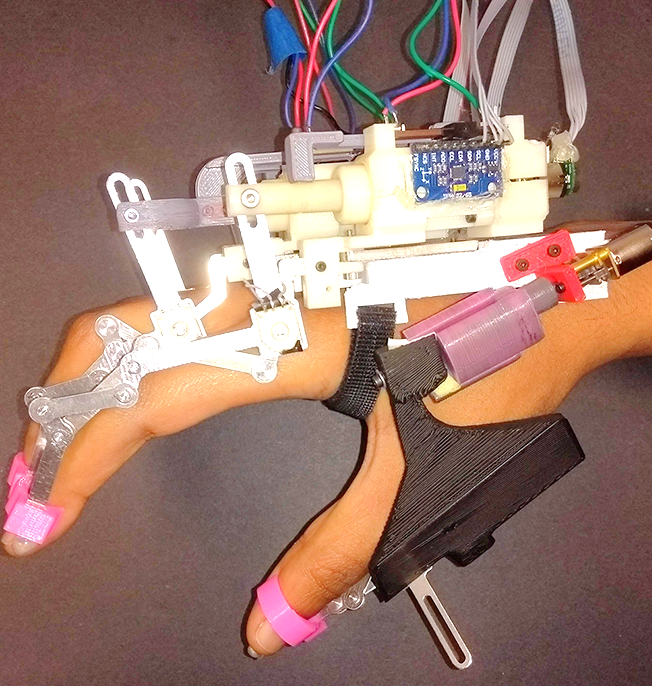}}
		& 
		
		Rehabilitative robotic glove \cite{Vanteddu2020}:
		
		- 1DoF rigid links paired with SEA
		
		- Helps paralysis patients with stable grasping objects. 
		
		- Uses force and moment equilibrium to prevent object deformation.
		\\ \vspace{1mm}
		
		Soft lower-extremity robotic exosuit \cite{Wehner2013}:
		
		- Do not restrict user's mobility. 
		
		- Uses McKibben style actuators.
		
		- Triangulated inextensible web.
		& \centering \raisebox{-\totalheight}{\includegraphics[scale=0.15]{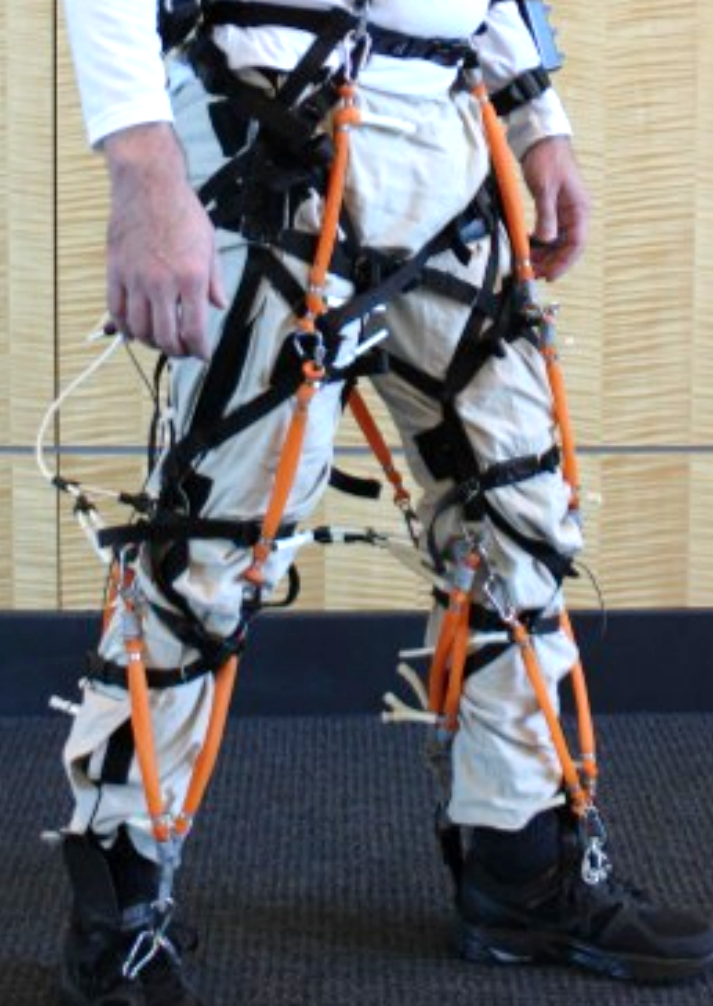}}
		& \centering \raisebox{-\totalheight}{\includegraphics[scale=0.25]{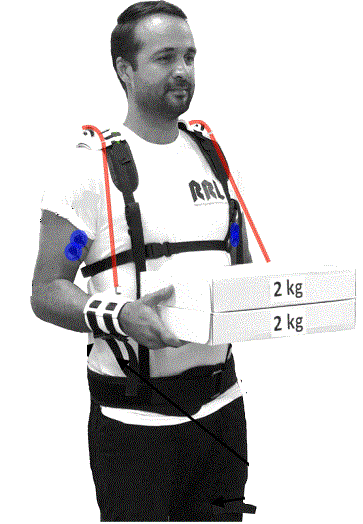}} 
		& \vspace{1mm}
		
				The TSA ExoSuit by Hosseini et al. \cite{Hosseini2020}:
		
		- sEMG-driven
		
		- Utilises Twisted String Actuators (TSAs)
		
		- Light weight/ 1650 g total weight
		
		- Up to 220\% and 110\% muscular activity compensation for single and dual arm assist.
		\\
		\hline
	\end{tabular}
\end{table*}

\subsection{Semi-passive exoskeletons}
One of the drawbacks of passive devices is their fix assist ratio defined by the stiffness of the elastic parts, limiting their adaptability to different situations. Semi-passive devices use low-power actuators to control the assistive ratio while maintaining low complexity and being lightweight. Jamsek et al. developed a spinal exoskeleton that uses electrical clutches to engage the elastic part during lifting and disengage during other tasks \cite{Jamsek2020}. H-PULSE uses an "active tuning mechanism" to automatically adjust the assistive torque using a servomotor, a spindle drive and a hierarchy control system \cite{Grazi2020a}. Walsh et al. optimised a quasi-passive design of a leg exoskeleton with linear springs in the ankle and hip and variable damper in the knee \cite{Walsh2007}. It finds the gait phase through knee angle and force and adjusts the variable-damping mechanism to help with negative mechanical power. The system decreased the metabolic cost by 11\% when fully engaged compared to a zero-impedance design.

Regarding ground-level walking, the knee shows the highest springy behaviour compared to other lower limb joints \cite{Shamaei2014}. To get the best results with minimum effort, researchers from Yale University worked on a quasi-passive knee exoskeleton that uses a finite-state machine to control the engagement \cite{Shamaei2014}. It determines the gait phase based on the sensors to engage the assistance spring during the weight acceptance phase and disengages it otherwise. Table \ref{tbl:semi} describes principles of some examples of semi-passive devices in the literature.

\subsection{Active exoskeletons}
\looseness=-1 While passive assistance can significantly help in preventing muscle fatigue in repetitive work, it may not be sufficient for some applications like power augmentation \cite{Zoss2006}, rehabilitation \cite{Lyu2019} or for weak and impaired users \cite{Sankai2010b}. Active exoskeletons use various sensors paired with complex computing units to control the joint movements in real time. They are mechanical structures whose joints are aligned with the user \cite{Yang2008}. The idea is to combine the robot's strength with human intelligence \cite{Kazerooni2005,Yang2008}.


Exoskeletons require a different control scheme compared to other types of robotics \cite{Yang2008}; The generated motion must follow the user's intention with minimum error \cite{Bogue2015}. Thus, an elaborate "human-in-the-loop" control system is required to synchronise the robot's movement with the user's intention based on HMI input. The major ordeal is predicting the user's intent and reacting to it in real-time \cite{Lenzi2012}. The intention will act as the system's input, and the output will be the exoskeleton's movement. The human will sense this movement as motion feedback which makes the user optimise their intention \cite{Yang2008}. Furthermore, human learning ability and motor adaptation play a crucial role in this synchronisation \cite{Gordon2007, Gordon2013}. This will raise the concern for cross adaptation in which the exoskeleton and the user shall adapt themselves to each other \cite{Ronsse2011}. One approach here is to calculate the joint torque to do a task and help the user with a pre-determined fraction of that torque, called assist ratio \cite{Kong2009}.

There are different ways to control the exoskeletons and HRI. Berkeley Lower Extremity Exoskeleton (BLEEX) minimises the force interaction between the human and robot by shadowing the user’s movement \cite{Kazerooni2005}. However, the problem here is to meet the sweet spot for the system’s sensitivity to the interaction force. This is essential to balance the system’s response time and its robustness. High sensitivity can cause losing stability as well as decreasing the robot’s precision to its dynamic model \cite{Kazerooni2005}. In BLEEX, the gait cycle has been divided into load support and swing phase, where position control and positive feedback control systems have been applied to them, respectively \cite{Kazerooni2006,Kazerooni2005}. The controller does not need direct input from HMI in the sensitivity amplification and only uses built-in accelerometers and encoders.

Another popular method is getting commands through biological signals. The most common signals are EMG, EEG, Electrooculogram (EOG), Electrocorticogram (ECoG) and Magnetoencephalography (MEG).  Electromyogram intent detection is categorised into classification \cite{Fajardo2021} and regression \cite{Bi2019} models. The system can estimate the joint torque and help with a fraction \cite{Ronsse2011}. The EMG signal starts 20-80ms before muscle activation \cite{Norman1979} making it useful for real-time processing \cite{Lenzi2011}. Another advantage of EMG-based control methods is that they can compensate for the patient’s inability to produce adequate joint torque \cite{Peternel2016}. In other words, as long as the patient can produce the neuro-musculoskeletal signal, in theory, they should be able to control the assistive device. However, it cannot be extended to people with motor injury \cite{Lyu2019}. There is a correlation between EMG signals and muscle force or intended joint torque\cite{Buchanan2004}. The output depends on the instrument, user’s anatomy, sensor placement, muscle crosstalk, and other noise sources.

\looseness=-1 One of the most known myoelectrically controlled exoskeletons are the HAL series. The development of HAL started in 1992 by Sankai et al. \cite{Sankai2010b}. HAL-1 was designed to help the user with the joint torque while walking, utilising DC motors and ball screws \cite{Sankai2010b}. HAL-3 used myoelectricity to implement the assist torque and sequence control \cite{Kawamoto2003}. Phase Sequence control divides walking into different phases for assistance based on muscle force conditions. Floor reaction forces can help with the transition between phases. Assist torque control, on the other hand, utilises a feedback controller to maintain the assistance ratio. "Cybernetic Voluntary" and "Autonomous Control" are the two complementary parts of the system. The first one helps with physical support based on the user’s intention extracted from the biosignal. The latter utilises different input types to control the system seamlessly, e.g. position-based and force-based inputs. In HAL-5, the shift of the centre of gravity helps to determine when to assist with walking \cite{Sankai2010b}.

\begin{figure}[t!]
	\centering
	\includegraphics[scale=0.6]{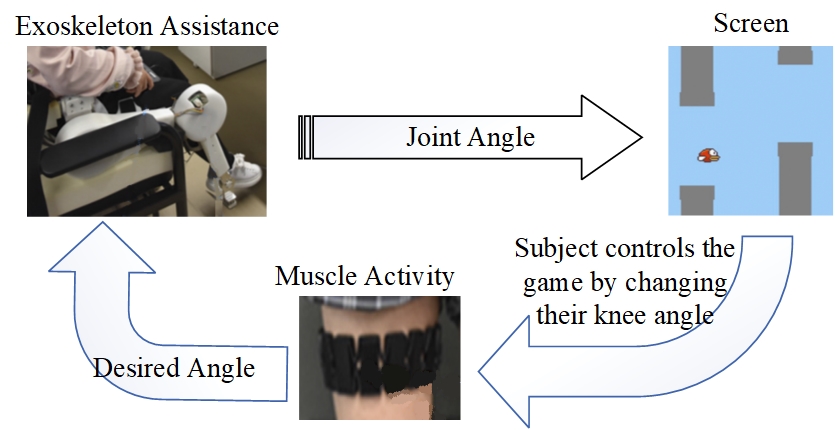}
	\caption{\looseness=-1 Closed-loop control system of rehabilitation system developed by Lyu et al. \cite{Lyu2019}.}\label{game}
\end{figure}

The active exoskeleton's design differs based on its application; performance augmentation \cite{Kazerooni2005, Lee2018, Kim2019a}, rehabilitation \cite{Yap2015, Lyu2019} and individuals with disabilities \cite{Young2017, Sankai2010b}. Assistance-as-needed knee exoskeleton developed by Lyu et al. is an example of robotic therapy \cite{Lyu2019}. It connects the knee exoskeleton to a screen game in which the subject controls the game by their knee angle and muscle activity as in Fig. \ref{game}. The inventors used AI agents based on a modular pipeline and deep Q-network to help with assistance as needed and enhance the exoskeleton's autonomous control.

An example of a soft active exoskeleton is the autonomous multi-joint soft exosuit developed by Wyss Institute \cite{Lee2018}. It helps the wearer with the plantarflexion and hip flexion/extension using iterative force-based position control and Bowden cables. By optimising control parameters for each individual, up to 15\% reduction in metabolic cost compared to not wearing the suit and 22\% compared to wearing the unpowered suit can be achieved. In another version of Wyss exosuit, only hip assistance was targeted to reduce the equivalent metabolic cost of 7.4 and 5.5 Kg of payload for walking and running, respectively \cite{Kim2019a}. Table \ref{tbl:active} highlights some famous active exoskeletons in the literature.

\section{Discussion}
Human assistive technologies, i.e. exoskeletons, need multiple areas of science to work together to effectively and efficiently perform their mission, considering the human body biomechanics \cite{Pons2008}. These devices face many challenges to become applicable in different situations and be adapted in various sectors. These challenges range from the mechanical design and energy source to an adequately high-level cognition and control system.

One of the fundamental fields in this area is the study and understanding the human motion kinetics and kinematics. Human limbs are complex systems, and their modelling differs for different joints and individuals. This is essential as these devices must be anthropomorphic and adapt to the limb's characteristics and motion. Another limitation is how these devices interact with the human body, which can be categorised into physical and cognitive interaction. Physical limitations include the imposed weight or pressure, motion restrictions, and added impedance. The cognitive limitations are mainly prominent in predicting the user's intention, extracting commands, and controlling the robot in real-time.

The following sub-section suggests some guidelines for exoskeleton selection based on the criteria that have been discussed in this paper, followed by future directions and possible mitigation for the current limitations facing this technology.

\subsection{Exoskeleton criteria and selection guidelines}
A design selection criterion of an exoskeleton is highly dependent on its application. For instance, an industrial exo-system needs to be cost-efficient and designed to operate in a specific environment \cite{Dahmen2020}. At the same time, a power augmentation exoskeleton for military or rescue purposes shall be highly reliable to work in various situations \cite{Proud2022}. A rehabilitation exoskeleton might be designed to impede \cite{Lim2019} or assist \cite{Noronha2021} with the joint motion of a recovering patient in a clinical environment, while an elderly care system may help with task completion and status monitoring of the user on daily life \cite{Balli2018, Zhang2020a}.

Also, an exoskeleton's physical and cognitive design is influenced by the target limbs/ joints the robot is assisting. As discussed earlier, an exoskeleton should be able to help the user in the desired direction with precise timing. Furthermore, it should not restrict or impede the user's motion unless that is intended. In other words, other than transparency and impedance considerations in the design, the system's DoF should be similar to the human. Also, its joints and limbs must stay aligned with them, as misalignment can impose unnecessary forces on the user's body and cause discomfort and fatigue.

After considering the mentioned criteria above, the next step is to design the assisting approach to task performance, motor skill training, rehabilitation, augmenting capabilities, etc. One of the most common ways assistive devices help users is with muscle engagement or joint torque \cite{Choi2021}. Actuators shall activate in precise moments to help or impede the motion of a limb/ joint. Muscle activation and metabolic cost are useful tools to measure the exoskeleton's effectiveness or study external assistance's effect on human limbs.

Special requirements specific to the user, environment or application shall be considered. For instance, if the user has a neuromuscular weakness, the device shall not rely on motion detection to predict the intention \cite{Sankai2010b}. This survey discussed alternative approaches that use physiological signals instead of kinetic and kinematic data \cite{Kiguchi2007, 8239668}. Another example to address special requirements is using compressed air as an energy source for factory applications where limited battery life could be an issue \cite{Kobayashi2009}. Safety, comfort, reliability, age and gender compatibility, cost and appearance are other special requirements that must be considered before exoskeleton design selection.

Having the mentioned criteria, a physical design can be proposed with respect to the user's biomechanics. This design should consider weight, size, range, actuation, and other criteria discussed earlier. The physical structure of an exoskeleton can be either soft or hard. Hard exoskeletons use rigid limbs connected through joints or sliding mechanisms \cite{Sankai2010b, Zoss2006}. They can be designed to transfer the load either to the human body or directly to the ground. A significant limitation in hard exoskeletons is that they must work tightly around human limbs/joints' kinematics. Soft exoskeletons, on the other hand, can be more forgiving thanks to the nature of their design \cite{Lee2018}. However, they cannot transfer the load directly to the ground, making them unsuitable for some heavy lifting tasks. Also, they use the user's bones for compression forces, limiting the applicability for some users and applications.

\begin{figure}[t]
	\centering
	\includegraphics[scale=0.66]{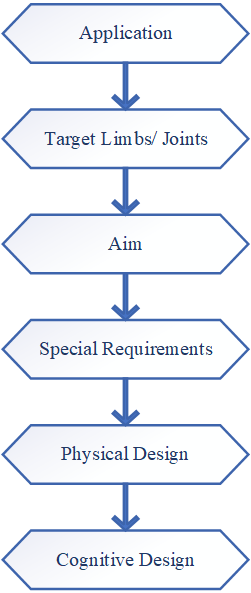}
	\caption{Exoskeleton selection guideline diagram.} \label{selection}
\end{figure}

Exoskeletons can be categorised based on their intended use, structural design or power consumption. One standard classification of these systems based on their actuation, control system and level of complexity is passive, semi-passive and active devices. Passive devices are the less complex, lightweight and economical solutions; They use an elastic element to save the motion energy and release it when required. Semi-passive exoskeletons use simple control components and light actuators to control the assistance ratio based on the situation and task. An active assistive device uses an elaborate cHRI system to monitor the user and environment's status and control the powered robot's joints in real-time \cite{Sankai2010b, Zoss2006}.

The final step in exoskeleton selection is the Cognitive design. cHRI defines the smartness of the system, whether autonomous or based on the user's intention. It starts by detecting the user and robot's state and predicting their intention \cite{Kazerooni2006}. This predicted intention can be the input to the low-level control system which controls the robot's actuators \cite{Kazerooni2005}. Regardless of whether or not the robot closes its feedback loop by monitoring the output, the human senses the action and will adjust their action, which will be sensed by the robot and closes the control loop \cite{Losey2018}.

For an intuitive control of an active exoskeleton, both the physical and cognitive HRI should be optimised. For instance, the mechanism should be lightweight; actuators and transmissions should be able to provide high torque and speed; and at the same time, do not add unwanted angular and linear momentum to the joint motion. An intent detection system is responsible for extracting the commands for intuitive control of the exoskeleton. This is one of the critical aspects of cognitive system design for exoskeletons. Intent interpretation is the act of mapping input signals from sensors to the desired intention. This can be achieved using either model-based or model-free approaches. With the increasing popularity of artificial intelligence (AI) and data-based techniques, AI model-free approaches have gained popularity. These systems can be developed to be more versatile and adaptable to different individuals, situations and tasks. Different methods for intent detection/interpretation have been proposed in the literature, from sensors combination and their positions to signal processing and intent prediction. Fig. \ref{selection} presents a guideline for exoskeleton selection for different purposes.

Several methods could be used to measure the effectiveness of the final design of an exoskeleton. An augmentation system could be measured against muscle torque reduction \cite{VandenBogert2003} or average maximum muscle load \cite{Li2018}, while the joint moment or angle can help with assessing a rehabilitation device\cite{Lyu2019}. Muscle activity \cite{Hull2020} and metabolic cost \cite{Collins2015} are some other widely used measures by researchers.

\subsection{Future directions to mitigate the limitations}
Like any other multidisciplinary area, improvements in other technologies can affect exoskeleton design and mitigate the current limitations in exoskeleton design and application. Here are a few areas that could improve with technology:

\begin{itemize}
\item Improvements in composite and nano-materials could lead to the design of structural mechanisms for exoskeletons with higher strength, lighter weight, smaller size and comfortable appearance and reduced impedance if required. It is also possible that researchers may try to combine hard and soft designs to use the benefit of both.

\item Developments of artificial muscles with the embedment of piezoelectric materials in them could enable versatile and responsive actuation with real-time force and position control. This changes the physical design and transforms exoskeletons into robots operating like the human body at the actuation level. They can reduce weight and eliminate the need for housing and transmission required by some electric motors.

\item Advances in sensors can lead to recording clearer signals with less noise-to-signal ratio from more challenging spots, e.g. hidden muscles. This can help to estimate the intent with higher accuracy and less uncertainty.

\item Advancements in batteries can help exoskeletons reduce weight and operate for an additional time.

\item Wireless energy transfer technologies like energy beamforming could be a potential candidate to mitigate the weight, transparency and range issues with exoskeletons. However, these technologies are mainly being tested on autonomous devices like \cite{Guo2021}, and human safety and approaches to prevent tissue harm from radiation shall also be investigated.

\item The fifth generation of wireless communication and the Internet of Things allows for streaming the processing work to the cloud, enabling computationally heavy algorithms on these portable devices without incurring additional cost, weight and power consumption.
\end{itemize}

\section{Conclusions}
This study surveyed the exoskeleton technology from both physical and cognitive Human-Robot Interaction perspectives. Different physical design criteria and limitations like human biomechanics, mechanism design, choice of inputs and outputs for the robot, and energy source have been covered. We discussed that the desired system needs to be anthropomorphic and remain aligned with the user's limbs during the motion/ task while not imposing unwanted restrictions or inertia on their body. What is more, it shall be quick and responsive.

\looseness=-1 Other than these physical criteria and limitations, we also discussed those on the cognitive level. The cognitive design starts from intent detection on the sensor level and signal processing, and extends to high- and low-level control systems. The literature suggests numerous attempts to implement different methods and strategies to improve the cognition of these systems.

Then, we explored some of the most renowned exoskeletons by categorising them into passive, semi-passive and active systems. Their design, application, innovation and limitations have been discussed. And finally, in the discussion, we showed how special requirements of the user and application, target limb and assistance type could help with the design selection criteria and proposed some potential direction and technological mitigation that can change the application of these devices.






\bibliographystyle{IEEEtran}

\bibliography{library}

\vspace{-5mm}
\begin{IEEEbiography}[{\vspace{-11mm} \includegraphics[width=1in,height=1.25in,clip,keepaspectratio]{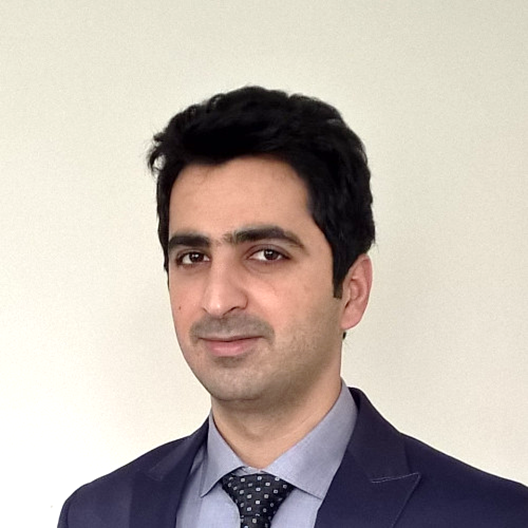}}]{Farhad Nazari}
received the B.Sc. degree in Mechanical Engineering from Isfahan University of Technology in 2013 and Masters of Mechanical Engineering from University of Tehran in 2017.
He is currently a PhD student in the Institute for Intelligent Systems Research and Innovation (IISRI), Deakin University. His research interests are human assistive devices and AI.
\end{IEEEbiography}
\vspace{-12mm}
\begin{IEEEbiography}[{\includegraphics[width=1in,height=1.25in,clip,keepaspectratio]{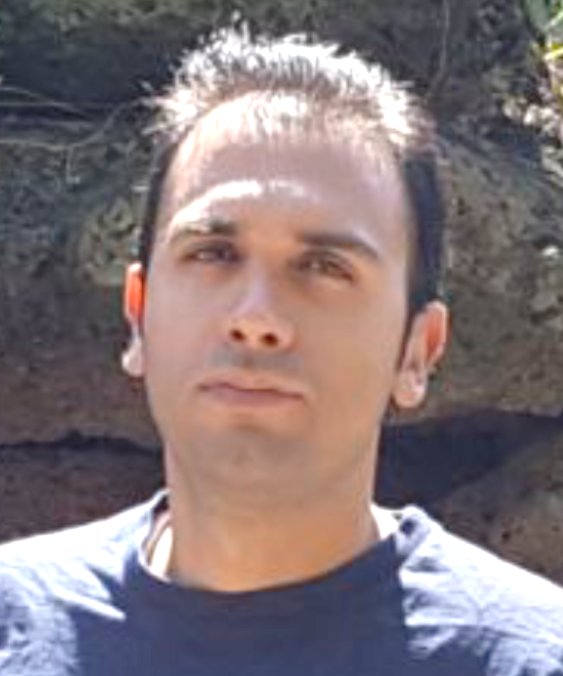}}]{Navid  Mohajer} received  the  B.Eng.  degree in  Mechanical  Engineering,  in  2009,  the  M.Sc. degree in Mechatronics Engineering from University of Tehran, Iran, in 2012, and the Ph.D. degree in Vehicle Dynamics and Mechanical Engineering form Deakin University, Australia, in 2017. He is currently a researcher within Deakin University,  Australia.  His  research  interests  include control  and  dynamic  of  Autonomous  Vehicles, mechanical design and analysis of complex system (kinematics and dynamics) and Multibody Systems (MBS).
\end{IEEEbiography}
\vspace{-8mm}
\begin{IEEEbiography}[{\includegraphics[width=1in,height=1.25in,clip,keepaspectratio]{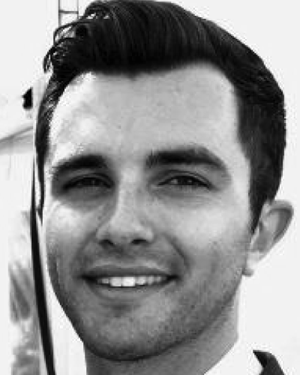}}]{Darius Nahavandi} received the Bachelor of Health Science and Ph.D. degrees from Deakin University, Geelong, VIC, Australia, in 2015 and 2018, respectively.
He is currently a Research Fellow with the Institute of Intelligent Systems Research and Innovation, Deakin University. Throughout the years, he has contributed to a range of projects involving the automotive, mining, and defense industries with a primary focus on innovation for human factors. His research consists of investigating human performance factors using advanced technologies both in the lab environment and industry settings.
\end{IEEEbiography}
\vspace{-8mm}
\begin{IEEEbiography}[{\includegraphics[width=1in,height=1.25in,clip,keepaspectratio]{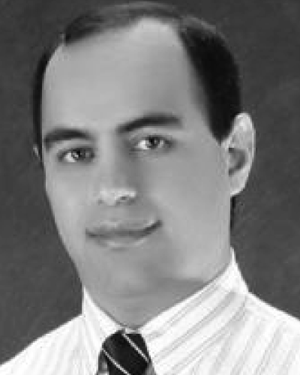}}]{Abbas Khosravi} is an associate professor with the Institute for Intelligent Systems Research and Innovation, Deakin University, Australia. He completed his PhD in machine learning at Deakin University in 2010. His broad research interests include artificial intelligence, deep learning, and uncertainty quantification. He is currently researching and applying probabilistic deep learning ideas and uncertainty-aware solutions in healthcare.
\end{IEEEbiography}
\vspace{-8mm}
\begin{IEEEbiography}[{\includegraphics[width=1in,height=1.25in,clip,keepaspectratio]{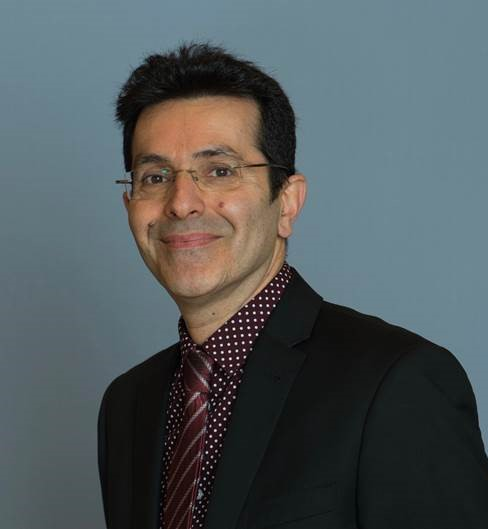}}]{Saeid Nahavandi} received a Ph.D. from Durham University, U.K. in 1991. He is an Alfred Deakin Professor, Pro Vice-Chancellor, Chair of Engineering, and the Director for the Institute for Intelligent Systems Research and Innovation at Deakin University. His research interests include modeling of complex systems, robotics and haptics. He has published over 800 scientific papers in various international journals and conferences. 
He is a Fellow of Engineers Australia (FIEAust), the Institution of Engineering and Technology (FIET) and Senior member of IEEE (SMIEEE).
He is the Co-Editor-in-Chief of the IEEE Systems Journal, Associate Editor of the IEEE/ASME Transactions on Mechatronics, Associate Editor of the IEEE Transactions on Systems, Man and Cybernetics: Systems, and IEEE Press Editorial Board member.
\end{IEEEbiography}\vfill
\end{document}